\documentclass[letterpaper, 10 pt, conference]{ieeeconf}  

\IEEEoverridecommandlockouts                              

\usepackage{algorithmic}
\usepackage{array}
\usepackage[caption=false,font=small,labelfont=sf,textfont=sf]{subfig}
\usepackage{textcomp}
\usepackage{stfloats}
\usepackage{url}
\usepackage{verbatim}
\usepackage{graphicx}

\usepackage{subfloat}
\usepackage{times}
\usepackage{amsmath,amssymb,amsopn,amstext,amsfonts}
\usepackage{cancel}
\usepackage[space]{cite}
\usepackage{pdfsync}
\usepackage{balance}
\usepackage{color}
\usepackage{mathtools}
\usepackage[ruled,vlined]{algorithm2e}
\usepackage{bm}

\usepackage{diagbox}
\usepackage{float}
\usepackage{epstopdf}
\usepackage{pifont}
\usepackage{fixltx2e}
\usepackage{amsmath}
\usepackage{multirow}
\usepackage{booktabs}
\usepackage{svg}
\usepackage{threeparttable}
\usepackage{caption}
\usepackage{lipsum}
\usepackage[linkcolor=black,citecolor=black,urlcolor=black,colorlinks=true]{hyperref}
\usepackage{makecell}

\begin{document}
\captionsetup{font={small}}
\title{\LARGE \bf
Environment Transformer and Policy Optimization for Model-Based Offline Reinforcement Learning
}

\author{Pengqin Wang$^{1,2}$, Meixin Zhu$^{1,2,3, \dag}$, and Shaojie Shen$^{1}$
\thanks{$^{1}$The Hong Kong University of Science and Technology, Hong Kong, 999077, China (email: pwangas@connect.ust.hk; eeshaojie@ust.hk).
        {\tt\small }}%
\thanks{$^{2}$The Hong Kong University of Science and Technology (Guangzhou), Guangzhou, 511400, China (email: meixin@ust.hk).
        {\tt\small }}%
\thanks{$^{3}$Guangdong Provincial Key Lab of Integrated Communication, Sensing and Computation for Ubiquitous Internet of Things, Guangzhou, 511400, China (email: meixin@ust.hk).
        {\tt\small }}%
\thanks{{\dag} Corresponding author}
}

\maketitle
\thispagestyle{empty}
\pagestyle{empty}

\begin{abstract}

Interacting with the actual environment to acquire data is often costly and time-consuming in robotic tasks. Model-based offline reinforcement learning (RL) provides a feasible solution. On the one hand, it eliminates the requirements of interaction with the actual environment. On the other hand, it learns the transition dynamics and reward function from the offline datasets and generates simulated rollouts to accelerate training. Previous model-based offline RL methods adopt probabilistic ensemble neural networks (NN) to model aleatoric uncertainty and epistemic uncertainty. However, this results in an exponential increase in training time and computing resource requirements. Furthermore, these methods are easily disturbed by the accumulative errors of the environment dynamics models when simulating long-term rollouts. To solve the above problems, we propose an uncertainty-aware sequence modeling architecture called Environment Transformer. It models the probability distribution of the environment dynamics and reward function to capture aleatoric uncertainty and treats epistemic uncertainty as a learnable noise parameter. Benefiting from the accurate modeling of the transition dynamics and reward function, Environment Transformer can be combined with arbitrary planning, dynamics programming, or policy optimization algorithms for offline RL. In this case, we perform Conservative Q-Learning (CQL) to learn a conservative Q-function. Through simulation experiments, we demonstrate that our method achieves or exceeds state-of-the-art performance in widely studied offline RL benchmarks. Moreover, we show that Environment Transformer's simulated rollout quality, sample efficiency, and long-term rollout simulation capability are superior to those of previous model-based offline RL methods.

\end{abstract}

\section{Introduction}

Deep reinforcement learning (RL) has obtained significant achievements in a variety of domains by utilizing a great deal of interactions with the environment\cite{silver2016mastering, vinyals2019grandmaster}.
However, due to the high expense of online data collection, the trial-and-error method is typically impractical in numerous real-world scenarios such as autonomous driving, robot manipulation, and aerial vehicles\cite{lee2022spendrobottime,ral2022,jin2022,zhao2022, gao2020teach, zhou2023racer}. 
Offline RL aims to solve the problem of learning a policy completely from a fixed batch of data without interacting with the environment \cite{SAC,BCQ,BRL}. This provides an appealing paradigm for a wide range of applications where there exist large and diverse pre-recorded datasets. Recent studies have shown that model-based RL, which learns the transition dynamics from the batch of data, demonstrates better generalization capability in dealing with offline RL tasks \cite{ROMI,kidambi2020morel,MOPO,COMBO,RAMBO}. 
Previous model-based offline RL approaches \cite{PETS,ROMI,kidambi2020morel,MOPO,COMBO,RAMBO} use probabilistic ensemble NN to capture both aleatoric uncertainty (inherent system stochasticity) and epistemic uncertainty (subjective uncertainty, due to limited data). Nevertheless, this leads to a significant growth in the duration of training and the need for computational resources. Furthermore, the effectiveness of these approaches might be significantly affected by the accumulative errors of the environment dynamics models when simulating long-term rollouts.

\begin{figure}[t]
	\begin{center}
		\includegraphics[width=1.0\columnwidth]{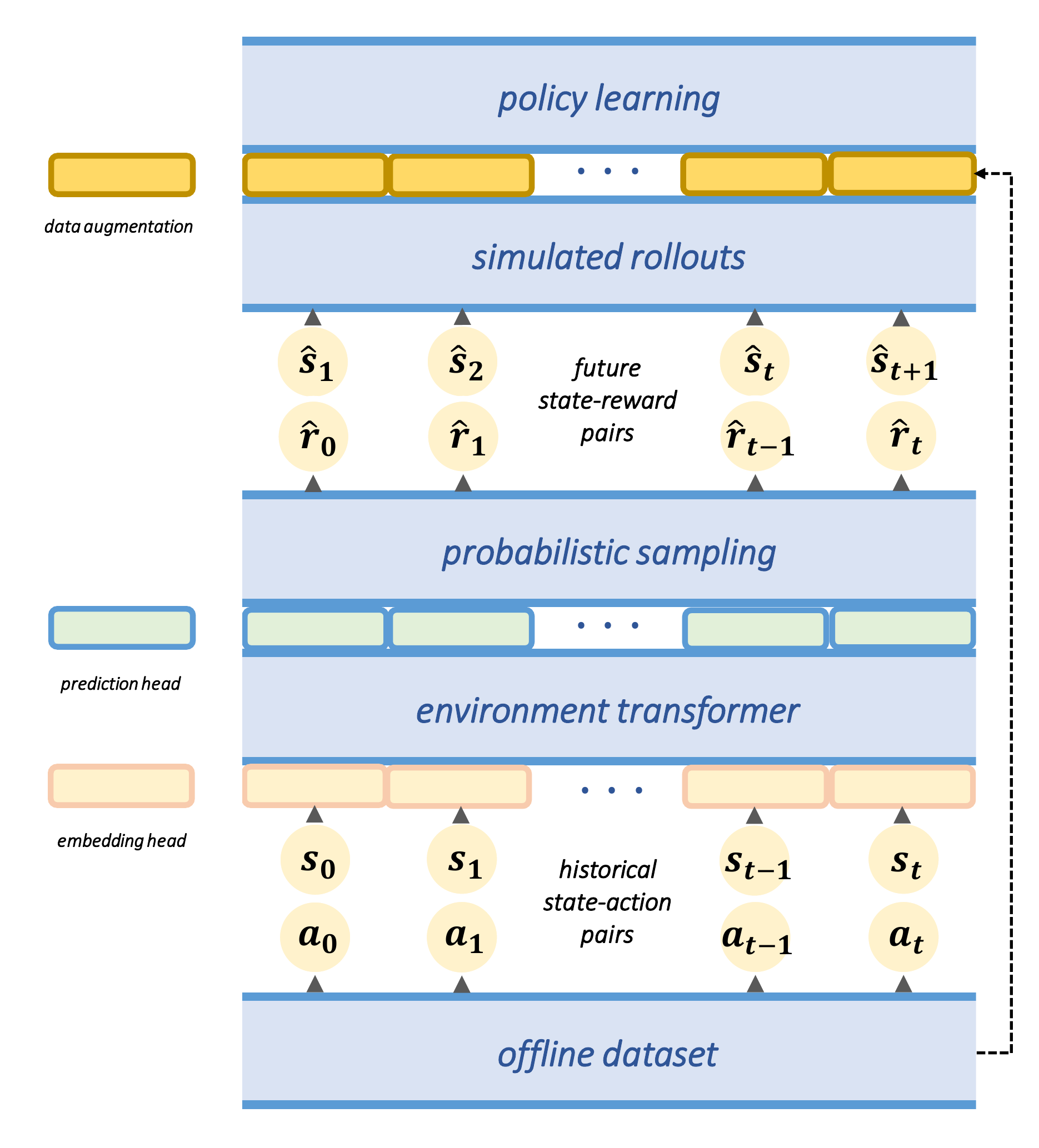}
	\end{center}
	\caption{
		\label{fig:algorithm}
		The framework of the proposed Environment Transformer.
	}
	\vspace{-0.8cm}
\end{figure}

To overcome the above issues, we consider the accurate modeling of the transition dynamics and reward function as a sequence-to-sequence task. We propose an uncertainty-aware sequence modeling architecture called Environment Transformer. It models the probability distribution of the environment dynamics and reward function to capture aleatoric uncertainty and treats epistemic uncertainty as a learnable noise parameter.
The state-action pairs are sampled from the offline dataset to predict the probability distribution of future state-reward pairs using a causal self-attention mask \cite{Attention,GPT}.
Furthermore, Environment Transformer can be combined with arbitrary planning, dynamics programming, or policy learning algorithms for offline RL, thanks to its accurate modeling of the transition dynamics and reward function. In this instance, we conduct Conservative Q-Learning (CQL) \cite{CQL} to learn a conservative Q-function, as shown in Fig. \ref{fig:algorithm}.

We compare the final performance of our proposed method with both model-based and model-free state-of-the-art (SOTA) offline RL methods in widely studied robot continuous control benchmarks \cite{mujoco, mujoco2}. Through simulation experiments, we show that our method achieves the highest score in 9 of 20 tasks and similar performance to SOTA in 8 tasks.
In addition, we demonstrate that the simulated rollout quality, sample efficiency, and long-sequence simulation error of Environment Transformer are superior to those of previous model-based offline RL methods.

We summarize the contributions of this paper as follows.

\begin{itemize}
	\item We propose Environment Transformer, an uncertainty-aware sequence modeling architecture. It models the probability distribution of the environment dynamics and reward function to account for aleatoric uncertainty and considers epistemic uncertainty as a learnable noise parameter. CQL is conducted based on Environment Transformer for offline RL tasks.
	\item Simulation experiments are performed on widely studied offline RL benchmarks. The experiment results show that our method achieves the highest score in 9 of 20 tasks and similar performance to SOTA in 8 tasks. Moreover, we demonstrate that Environment Transformer's simulated rollout quality, sample efficiency, and long-sequence simulation error are superior to those of previous model-based offline RL methods. Therefore, it provides a low-cost, high-efficiency data acquisition method for training real-world robots, with a wide range of application prospects.
\end{itemize}

\section{Related Work}

\subsection{Offline RL}

The research of Offline RL addresses the challenge of learning a policy from a fixed dataset instead of interacting with the environment. The major issue of Offline RL is extrapolation error \cite{BCQ}, which is a generalization error in estimating value functions caused by out-of-distribution actions. One potential approach to solving this problem is to integrate a pessimistic bias for unseen actions into the Q-function. Kumar et al. \cite{CQL} propose to learn a conservative Q-function such that the expected value of a policy under the Q-function lower-bounds its true value. Kostrikov et al. \cite{IQL} treat the state value function as a random variable whose randomness is determined by the action, and use the upper expectile to estimate the value of the best actions. An alternative approach is to restrict the policy optimization to remain close to the original data samples. Wang et al. \cite{CRR} utilize a form of critic-regularized regression to learn policies from data. Nair et al. \cite{AWAC} combine sample efficient dynamic programming with maximum likelihood policy updates. Wu et al. \cite{BRAC} propose a framework to generalize previous approaches to solve the offline RL problem by regularizing the behavior policy. Zhou et al. \cite{PLAS} propose to learn the policy in the latent action space, which constrains the policy to select actions within the support of the dataset. Xu et al. \cite{SBAC} construct a novel offline-applicable policy learning goal that corresponds to the advantage function value of the behavior policy, multiplied by a state-marginal density ratio. Fujimoto et al. \cite{TD3+BC} include a behavior cloning component in the policy update of an online RL algorithm and normalize the data to push the policy towards favoring actions in the dataset. 

\subsection{Dynamics Modeling}

There exists a wealth of prior works to learn the environmental dynamics. Sutton \cite{Dyna} proposes an architecture that maintains a  dynamics model of the agent's transitions. Deisenroth et al. \cite{PILCO} model the environment dynamics as Gaussian processes and incorporate model uncertainty into long-term planning. Levine et al. \cite{Guided} leverage local linear models to represent the environment dynamics. Chua et al. \cite{PETS} integrate deep network dynamics models that account for uncertainty with sampling-based uncertainty propagation. Janner et al. \cite{MBPO,MOPO} use bootstrap ensembles of predictive models to capture aleatoric uncertainty and epistemic uncertainty of the environment dynamics. 

Recent breakthroughs in sequence modeling using deep neural networks have resulted in fast improvements in long-horizon prediction accuracy and model efficiency \cite{STS, Attention, GPT}. Chen et al. \cite{DT, TT} consider RL to be a sequence modeling problem, which aims to generate a series of actions to receive high rewards. In our approach, environment modeling is viewed as a sequence-to-sequence problem, which is required to predict the probability distribution for future state-reward pairs based on historical state-action pairs.

\section{Preliminaries}

\subsection{Transformer}

Transformers are proposed by \cite{Attention} as a framework for effectively modeling sequential data, which consists of stacked encoders and decoders. Both of them are based on attention mechanisms, where the $i$-th input token $x_i$ is embedded and mapped to key $k_i$, query $q_i$ and value $v_i$, and the $i$-th output token can be represented as: 

\begin{align}
z_i = \sum_{j=1}^n {\rm{softmax}} \left(\left\{\left \langle q_i,k_{j'} \right \rangle\right\}^n_{j'=1}\right)_j\cdot v_j.
\end{align}

The generative pre-training transformer (GPT) is proposed by \cite{GPT}, where the encoder-decoder architecture is changed to a causal self-attention mask without any encoders. In our practice, we adopt the GPT framework and develop Environment Transformer to obtain accurate predictions for environment dynamics and reward function.

\subsection{Conservative Q-Learning}

We consider a Markov decision process (MDP), defined by the tuple ($\mathcal S$, $\mathcal A$, $T$, $r$, $\gamma$). $\mathcal S$ and $\mathcal A$ represent state and action space. We consider state space and action space to be both continuous. The transition dynamics is denoted as $T(s^\prime|s, a)$, and the reward function is written as $r(s, a)$, and $\gamma \in (0, 1)$ represents the discount factor.

The goal of standard RL algorithms is to obtain the optimal policy $\pi^*$ such that the cumulative reward is maximized. The optimal policy $\pi^*$ can be written as:

\begin{align}
\pi^* = \mathop{\arg\max}\limits_{\pi} \mathbb{E}\big[\sum_{t=0}^\infty r\left(s_t,a_t\right)\big].
\end{align}

Q-learning methods maintain a Q-function $Q(s,a)$ that measures the discounted return based on the state $s$ and action $a$, under current policy $\pi$. Given current policy $\pi(a|s)$, the Bellman backup for obtaining the corresponding $Q$ function gives:
\begin{equation}
    \label{eq:bellmanbackup}
    {\mathcal{B}}^{{\pi}} Q(s,a):= r(s,a) + \gamma\mathbb{E}_{s^\prime\sim T(s^\prime|s, a), {a^\prime\sim\pi(a^\prime|s^\prime)}}[Q(s^\prime, a^\prime)].
\end{equation}

The optimal policy's Q-function satisfies the following Bellman optimal operator:
\begin{equation}
    \label{eq:bellmanoptimal}
    {\mathcal{B}}^{{*}}Q(s,a) := r(s,a) + \gamma\mathbb{E}_{s^\prime\sim T(s^\prime|s, a)}\left[\max_{a^\prime\in\mathcal{A}}Q(s^\prime, a^\prime)\right].
\end{equation}

Online interaction with the environment is impractical in offline RL settings. Only previously collected datasets $\mathcal{D}=\{s_t,a_t,r_t,s_{t+1},d_t\}_{t=1}^N$ are accessible, where $d$ is the terminal flag showing whether the episode is ended. CQL approach \cite{CQL} learns a conservative Q-function such that the expected value of a policy under this Q-function lower-bounds its true value, which gives the following iterative update for training:

\begin{equation}
    \label{eq:cql}
    \begin{aligned}
        \min_{Q} {\max_{\mu}}~  \alpha ~  (& \mathbb{E}_{s\sim\mathcal{D},a\sim\mu(a|s)}\left[Q(s,a)\right] - \\
        & \mathbb{E}_{s\sim\mathcal{D},a\sim{\hat{\pi}_\beta(a|s)}}\left[Q(s,a)\right]) + \\
        & \frac{1}{2}~ \mathbb{E}_{{s,a,s^\prime}\sim\mathcal{D}}\left[\left(Q(s, a) - {\hat{\mathcal{B}}}^{{\pi}_k} \hat{Q}^{k} (s, a) \right)^2 \right] + \\
        & {\mathcal{R}(\mu)},
    \end{aligned}
\end{equation}
where ${\hat{\pi}_\beta(a|s)}$ represents the data distribution and ${{\pi}_k}$ is the policy derived from the Q-function. $\mathcal{D}$ is the collected offline dataset augmented by the simulated rollouts and $\mathcal{R}$ is a regularizer.

\section{Environment Transformer}

\subsection{Environment Modeling}

We consider the environment dynamics as a Gaussian distribution with diagonal covariance to capture aleatoric uncertainty and treats epistemic uncertainty as a learnable noise parameter: 

\begin{align}
Pr\left({s}_{t+1}, {r}_t \big| s_{t}, a_{t}\right) = 
\mathcal{N}\left({\mu_{au}}\left(s_{t}, a_{t}\right), {\Sigma_{au}}\left(s_{t}, a_{t}\right)\right) + \epsilon_t,
\label{gaussianprocess}
\end{align}
where $(s_{t}, a_{t})$ denotes the state-action pairs at time step $t$. $\mu_{au}(s_{t}, a_{t})$ and $\Sigma_{au}(s_{t}, a_{t})$ represent the mean and covariance of the Gaussian distribution considering aleatoric uncertainty at time step $t$, respectively. $\epsilon_t$ is a learnable parameter for epistemic uncertainty at time step $t$.

\subsection{Training}

The training inputs for Environment Transformer are state-action pairs from the offline datasets containing thousands of trajectories. The $i$-th trajectory training input $\tau^i$ can be represented as:

\begin{align}
\tau^i = \left\{\left(s^i_t, a^i_t\right)\right\}^{T-1}_{t=0},
\end{align}
where $t$ denotes the timestep of state-action pairs, $i$ is training trajectory index and $T$ is the sequence length.

In practice, we consider the epistemic uncertainty parameter $\epsilon_t$ to be a sample from a Gaussian distribution with zero mean and learnable covariance conditioned on input state-action pairs. After the prediction and probabilistic sampling, the output of Environment Transformer $\phi^i$ can be written as: 

\begin{align}
& f^i_t \sim \mathcal{N}\left({\mu_{au}}\left(s^i_{t}, a^i_{t}\right), {\Sigma_{au}}\left(s^i_{t}, a^i_{t}\right)\right), \\
& \epsilon^i_t \sim \mathcal{N}\left(0, {\Sigma_{eu}}\left(s^i_{t}, a^i_{t}\right)\right), \\
& \phi^i = \left\{\left(f^i_t + \epsilon^i_t\right)\right\}^{T-1}_{t=0},
\end{align}
where $\Sigma_{eu}(s^i_{t}, a^i_{t})$ represents the learnable covariance of the Gaussian distribution considering epistemic uncertainty conditioned on the input state-action pair $(s^i_{t}, a^i_{t})$.
$\mu_{au}(s^i_{t}, a^i_{t})$ and $\Sigma_{au}(s^i_{t}, a^i_{t})$ represent the mean and covariance of the Gaussian distribution considering aleatoric uncertainty at time step $t$ of the $i$-th trajectory.

After obtaining the predicted future state-reward pairs for the $i$-th trajectory, we can calculate the mean square error (mse) loss between the predictions and ground-truth:

\begin{align}
loss = mse\left(\phi^i, \left\{\left({s}^i_{t+1}, {r}^i_t\right)\right\}^{T-1}_{t=0}\right).
\end{align}

We train Environment Transformer on 4 Nvidia RTX 3080 10GB and Intel(R) Xeon(R) Platinum 8255C CPU. We use Gemini, the heterogeneous memory space manager of Colossal-AI \cite{colossalai}, to train Environment Transformer. The hyperparameters are listed in Table \ref{tbl:hyperparameters}.

\begin{table}[ht]
\renewcommand\arraystretch{1.3}
\caption{Hyperparameters of Environment Transformer}
\begin{center}
\begin{small}
\begin{tabular}{ll}\hline
\textbf{Hyperparameter} & \textbf{Value}  \\\hline
Number of layers & $8$  \\ 
Number of attention heads    & $16$  \\
Embedding dimension    & $1024$  \\
Nonlinearity function & ReLU \\
Batch size   & $16$ \\
Sequence length  & $100$ \\
Dropout & $0.1$ \\
Learning rate & $10^{-4}$ \\
Weight decay & $10^{-4}$ \\
Optimizer & HybridAdam \\
HybridAdam eps & $10^{-4}$  \\\hline
\end{tabular}
\end{small}
\label{tbl:hyperparameters}
\end{center}
\end{table} 

\begin{figure*}[htbp]
	\centering

        \subfloat[Ant]{
		\includegraphics[width=0.4\columnwidth]{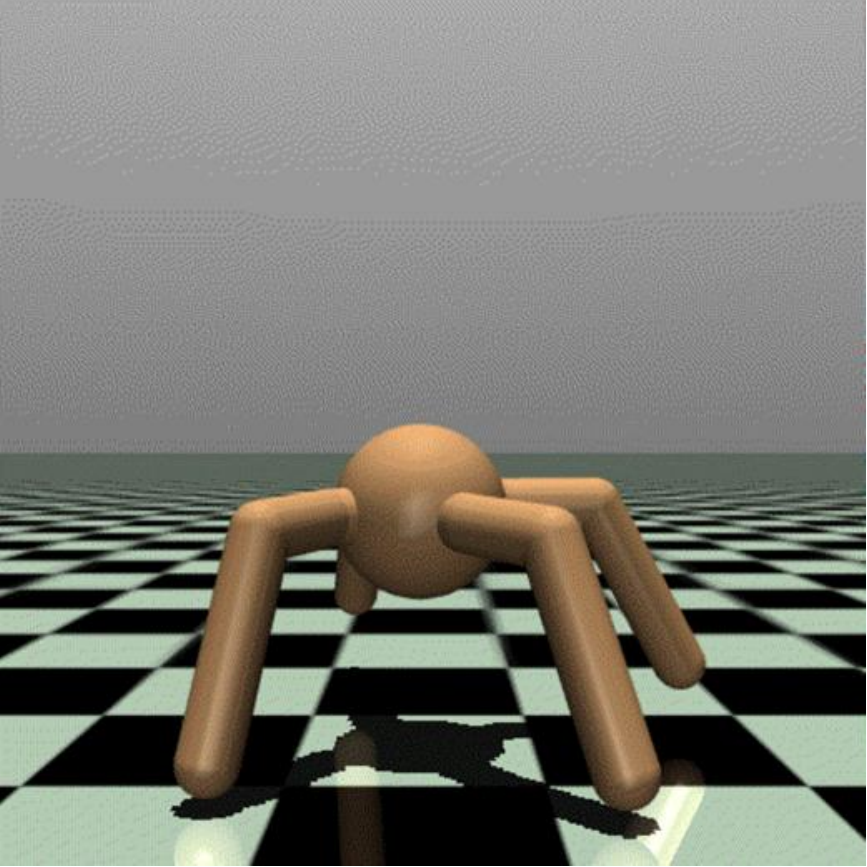}
		\label{fig:ant_scene}
	}	
	\subfloat[HalfCheetah]{
		\includegraphics[width=0.4\columnwidth]{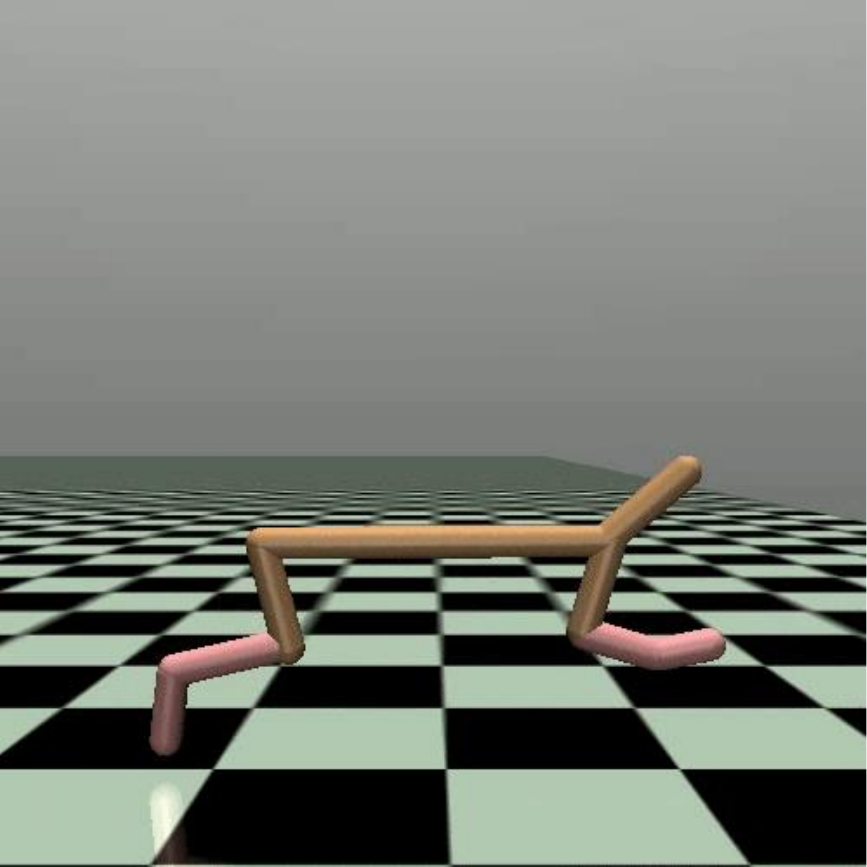}
		\label{fig:halfcheetah_scene}
	}
	\subfloat[Hopper]{
		\includegraphics[width=0.4\columnwidth]{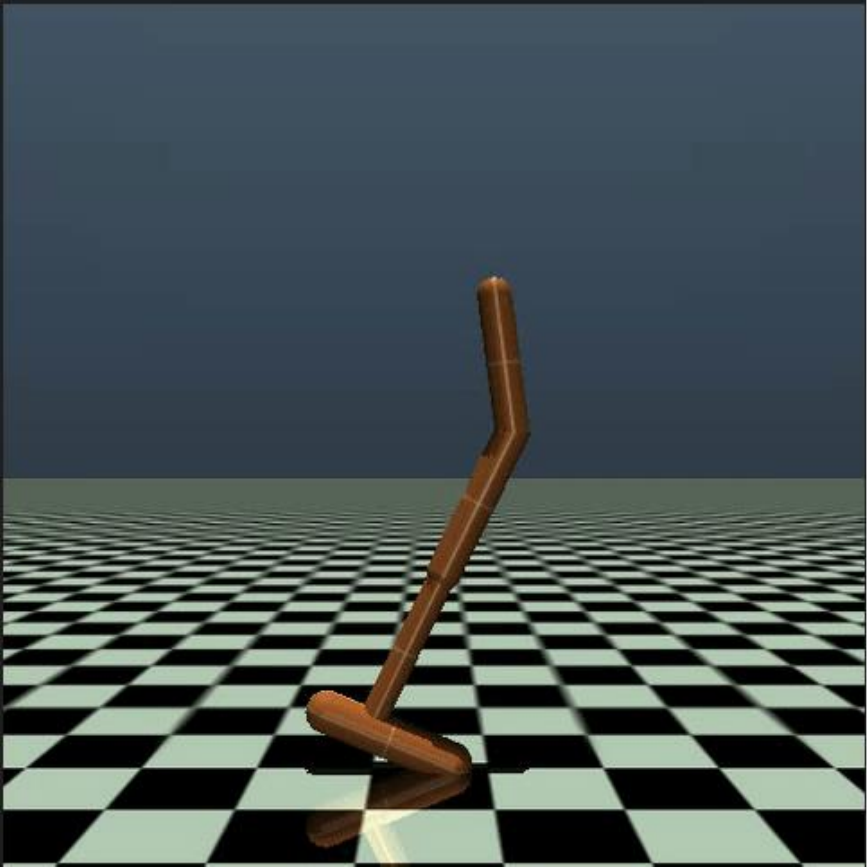}
		\label{fig:hopper_scene}
	}
	\subfloat[Walker2d]{
		\includegraphics[width=0.4\columnwidth]{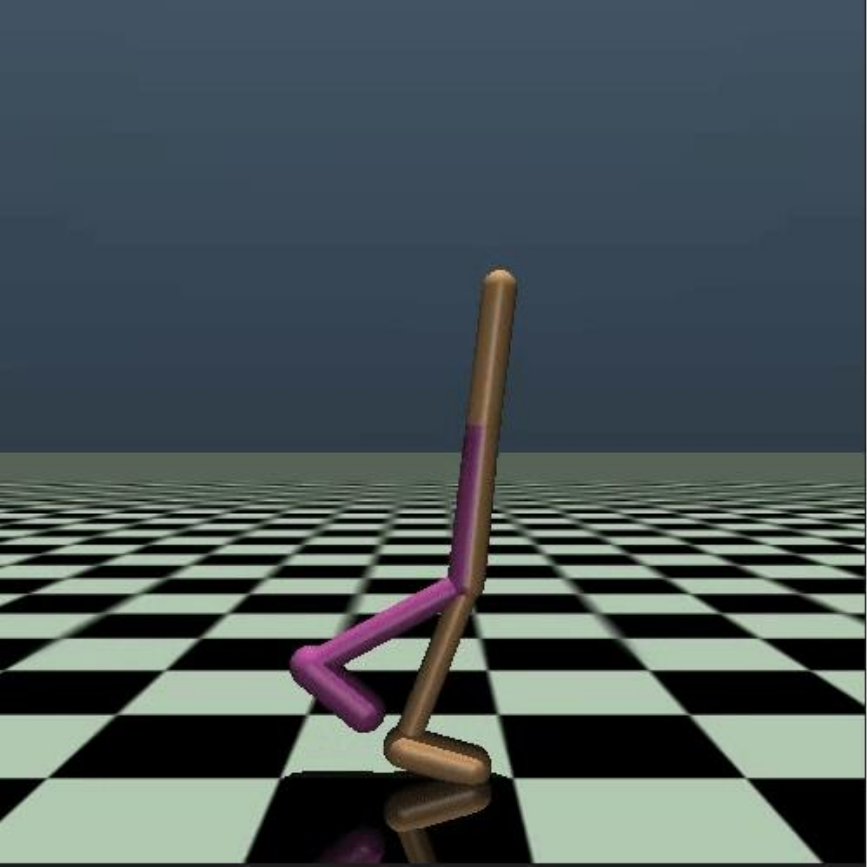}
		\label{fig:walker2d_scene}
	}
        
	\caption{
		The robot continuous control offline RL benchmarks including ant, halfcheetah, hopper and walker2d.
	}	
	\label{fig:mujoco}
\end{figure*}

\section{Experiments}

\begin{table*}[ht]
	\vspace{0.3cm}
	\renewcommand\arraystretch{1.4}
	\setlength{\tabcolsep}{0.5cm}
	\centering
	
	\begin{tabular}{ccccccccccc}
		\toprule
		\multicolumn{2}{c}{\textbf{\makecell{Environment}}}  & \multicolumn{2}{c}{\textbf{\makecell{Dataset}}}  & 
		\textbf{\makecell{Ours}}  & \textbf{\makecell{MOPO}}  & \textbf{\makecell{RAMBO}}  &
		\textbf{\makecell{COMBO}}  & \textbf{\makecell{CQL}}  & \textbf{\makecell{IQL}}  \\
		\hline
		
		\multicolumn{2}{c}{\makecell{}}  & \multicolumn{2}{c}{\makecell{Full-Replay}}  &  
		{80.8$\pm$0.4}  & {86.4}  & \textbf{89.0}  & -  & - & - \\
		
		\multicolumn{2}{c}{\makecell{}}  & \multicolumn{2}{c}{\makecell{Medium-Expert}}  &  
		\textbf{96.4}$\pm$0.6  & 63.3  & 79.3  & 90.0  & 62.4 & 86.7\\
		
		\multicolumn{2}{c}{\makecell{HalfCheetah}}  & \multicolumn{2}{c}{\makecell{Medium-Replay}}  &  
		{45.1$\pm$0.9}  & 53.1  & \textbf{67.0}  & 55.1  & 46.2 & 44.2\\

            \multicolumn{2}{c}{\makecell{}}  & \multicolumn{2}{c}{\makecell{Medium}}  &  
		{49.5$\pm$0.1}  & {42.3}  & \textbf{71.0}  & 54.2  & 44.4 & 47.4\\
  
		\multicolumn{2}{c}{\makecell{}}  & \multicolumn{2}{c}{\makecell{Random}}  &  
		{22.7$\pm$2.1}  & {35.4}  & {33.5}  & \textbf{38.8}  & 35.4 & -\\
		
		\hline
		
		\multicolumn{2}{c}{\makecell{}}  & \multicolumn{2}{c}{\makecell{Full-Replay}}  &  
		{106.9$\pm$0.9}  & \textbf{108.1}  & {107.6}  & -  & - & -\\
		
		\multicolumn{2}{c}{\makecell{}}  & \multicolumn{2}{c}{\makecell{Medium-Expert}}  &  
		{106.9$\pm$2.7}  & 23.7  & 89.5  & \textbf{111.1}  & 111.0 & 91.5\\
		
		\multicolumn{2}{c}{\makecell{Hopper}}  & \multicolumn{2}{c}{\makecell{Medium-Replay}}  &  
		{94.9$\pm$6.3}  & {67.5}  & \textbf{97.6}  & 73.1  & 48.6 & 94.7\\

            \multicolumn{2}{c}{\makecell{}}  & \multicolumn{2}{c}{\makecell{Medium}}  &  
		{67.3$\pm$6.5}  & {28.0}  & {91.2}  & \textbf{94.9}  & 86.6 & 66.3\\
  
		\multicolumn{2}{c}{\makecell{}}  & \multicolumn{2}{c}{\makecell{Random}}  &  
		{10.1$\pm$0.7}  & {11.7}  & {15.5}  & 17.9  & 10.8 & -\\
		
		\hline
		
		\multicolumn{2}{c}{\makecell{}}  & \multicolumn{2}{c}{\makecell{Full-Replay}}  &  
		\textbf{97.5}$\pm$1.2  & {56.9}  & {52.6}  & -  & - & -\\
		
		\multicolumn{2}{c}{\makecell{}}  & \multicolumn{2}{c}{\makecell{Medium-Expert}}  &  
		{84.7$\pm$2.4}  & 44.6  & 63.1  & 96.1  & 98.7 & \textbf{109.6}\\
		
		\multicolumn{2}{c}{\makecell{Walker2d}}  & \multicolumn{2}{c}{\makecell{Medium-Replay}}  &  
		\textbf{88.9}$\pm$4.6  & {39.0}  & {88.5}  & 56.0  & 32.6 & 73.9\\

            \multicolumn{2}{c}{\makecell{}}  & \multicolumn{2}{c}{\makecell{Medium}}  &  
		{76.6$\pm$3.1}  & {17.8}  & \textbf{89.1}  & 75.5  & 74.5 & 78.3\\
  
		\multicolumn{2}{c}{\makecell{}}  & \multicolumn{2}{c}{\makecell{Random}}  &  
		\textbf{14.2}$\pm$1.7  & {13.6}  & {0.2}  & 7.0  & 7.0 & -\\
		
		\hline
		
		\multicolumn{2}{c}{\makecell{}}  & \multicolumn{2}{c}{\makecell{Full-Replay}}  &  
		\textbf{138.4}$\pm$0.8  & {27.8}  & {119.3}  & -  & - & -\\
		
		\multicolumn{2}{c}{\makecell{}}  & \multicolumn{2}{c}{\makecell{Medium-Expert}}  &  
		\textbf{141.5}$\pm$2.3  & 26.8  & 95.7  & -  & - & -\\
		
		\multicolumn{2}{c}{\makecell{Ant}}  & \multicolumn{2}{c}{\makecell{Medium-Replay}}  &  
		\textbf{105.7}$\pm$1.9  & {28.4}  & 49.7  & -  & - & -\\

            \multicolumn{2}{c}{\makecell{}}  & \multicolumn{2}{c}{\makecell{Medium}}  &  
		\textbf{116.1}$\pm$0.5  & {20.4}  & {67.0}  & -  & - & -\\
  
		\multicolumn{2}{c}{\makecell{}}  & \multicolumn{2}{c}{\makecell{Random}}  &  
		\textbf{32.6}$\pm$8.2  & {15.8}  & {29.8}  & -  & - & -\\
		
		\hline
		
		\multicolumn{4}{c}{\makecell{Average}}  &  
		\textbf{78.8} & 40.5   & 69.8  & 64.1   & 54.9  & 77.0  \\

		\bottomrule
	\end{tabular}
	\caption{Comparisons on Offline RL Benchmarks}
	\label{tab:results}
        \vspace{-0.3cm}
\end{table*}

In this section, we design simulation experiments on widely studied offline RL benchmarks \cite{halfcheetah, hopperandwalker, ant} to evaluate the performance with both model-based and model-free SOTA offline RL algorithms. Then we compare the simulated rollout quality and sample efficiency with previous model-based offline RL methods based on probabilistic ensemble NN. Finally, we evaluate the model's capacity for long-term forecasting using both offline datasets and online environments.
The benchmarks include four environments (Ant, HalfCheetah, Hopper, and Walker2d) and five dataset types (full-replay, medium-expert, medium-replay, medium, and random), as shown in Fig. \ref{fig:mujoco}. The Ant is a 3D robot consisting of one torso with four legs, whose goal is to coordinate the four legs to move forward. The HalfCheetah is a two-dimensional robot. The objective is to apply torque to its joints so that the robot can run forward as quickly as possible. The Hopper is a one-legged, two-dimensional figure. Similarly, the goal is to make forward-moving jumps. The Walker2d is a two-dimensional figure with two legs, whose goal is to control the forward movement.

\subsection{Comparisons on Offline RL Benchmarks}

We evaluate the final performance after $1e6$ steps policy learning, in comparison with model-based approaches MOPO, COMBO and RAMBO\cite{MOPO,COMBO,RAMBO}, and model-free methods CQL and IQL\cite{CQL,IQL}. We report the mean and variance over three random seeds. The results are shown in Table \ref{tab:results}. Our approach is the strongest by a significant margin on all datasets in Ant environment with 111-dimensional observation space, which demonstrates that Environment Transformer can accurately model the aleatoric and epistemic uncertainty for complex environments. Our method achieves the strongest performance of full-replay, medium-replay, and random datasets in Walker2d environment because it is the environment with the second-highest complexity. However, our approach performs less well in Hopper environment. We analyze this because Hopper environment is relatively basic and the observation space is only 11 dimensional. As a result, previous probabilistic ensemble NN techniques are adequate for modeling it. On the whole, our method achieves the highest score in 9 of 20 tasks and similar performance to SOTA in 8 tasks.
In conclusion, our approach performs the best among both model-based and model-free SOTA offline RL algorithms.

\subsection{Simulated Rollout Quality and Sample Efficiency}

\begin{figure*}[htbp]
	\centering
	\vspace{-0.2cm}	
	\subfloat[ant-full-replay]{
		\includegraphics[width=0.5\columnwidth]{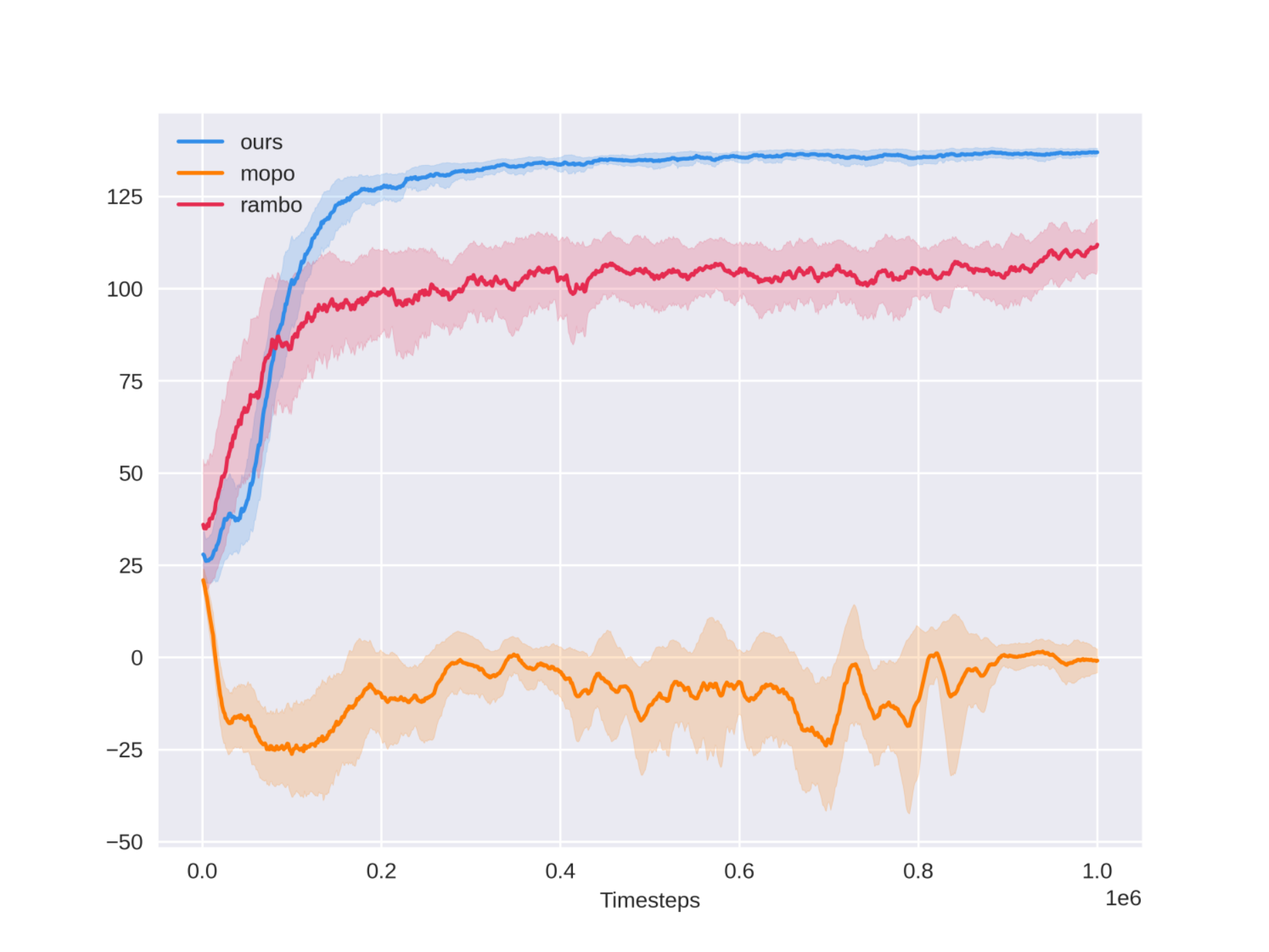}}
	\subfloat[ant-medium-expert]{
		\includegraphics[width=0.5\columnwidth]{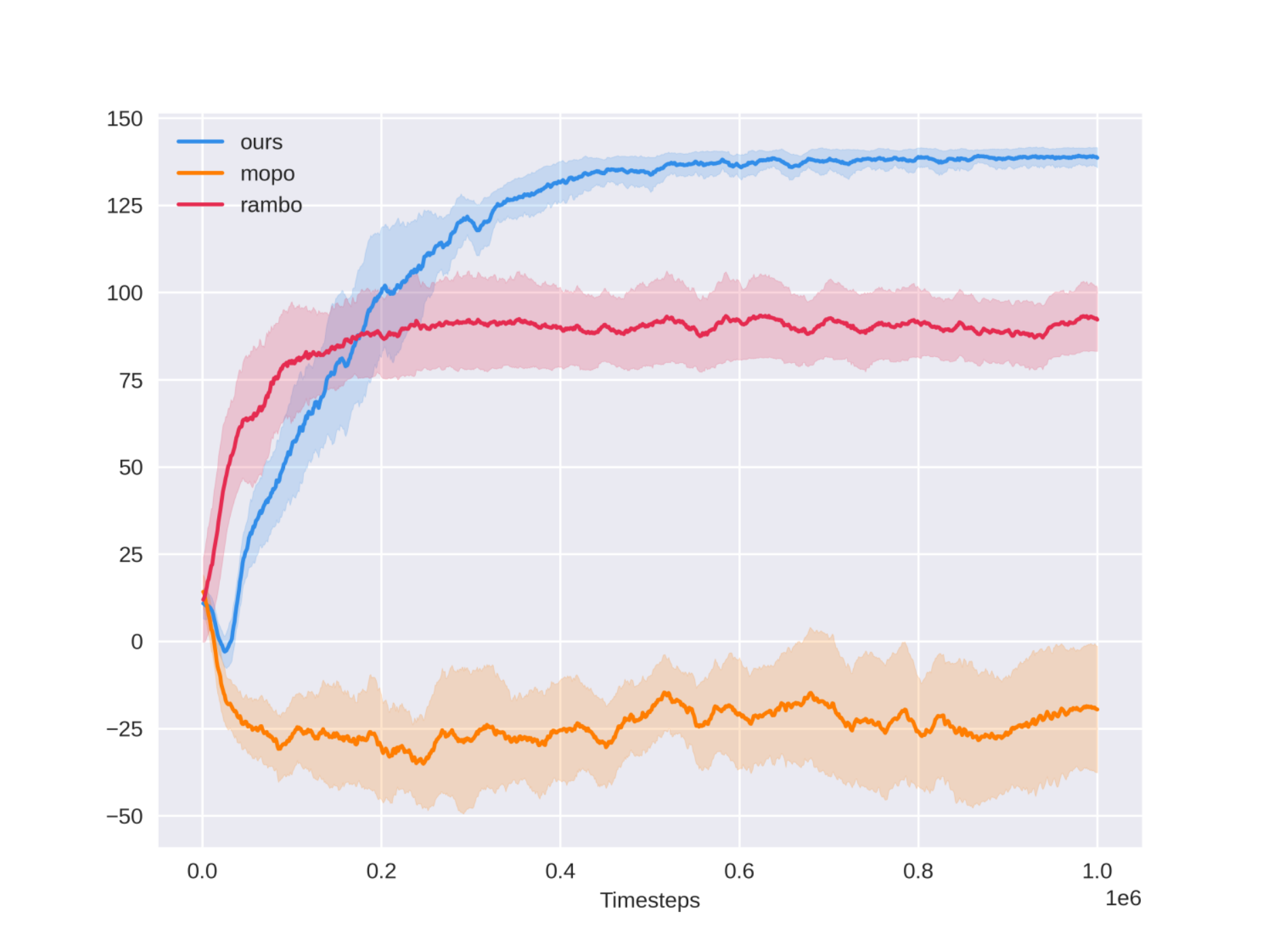}}
	\subfloat[ant-medium-replay]{
		\includegraphics[width=0.5\columnwidth]{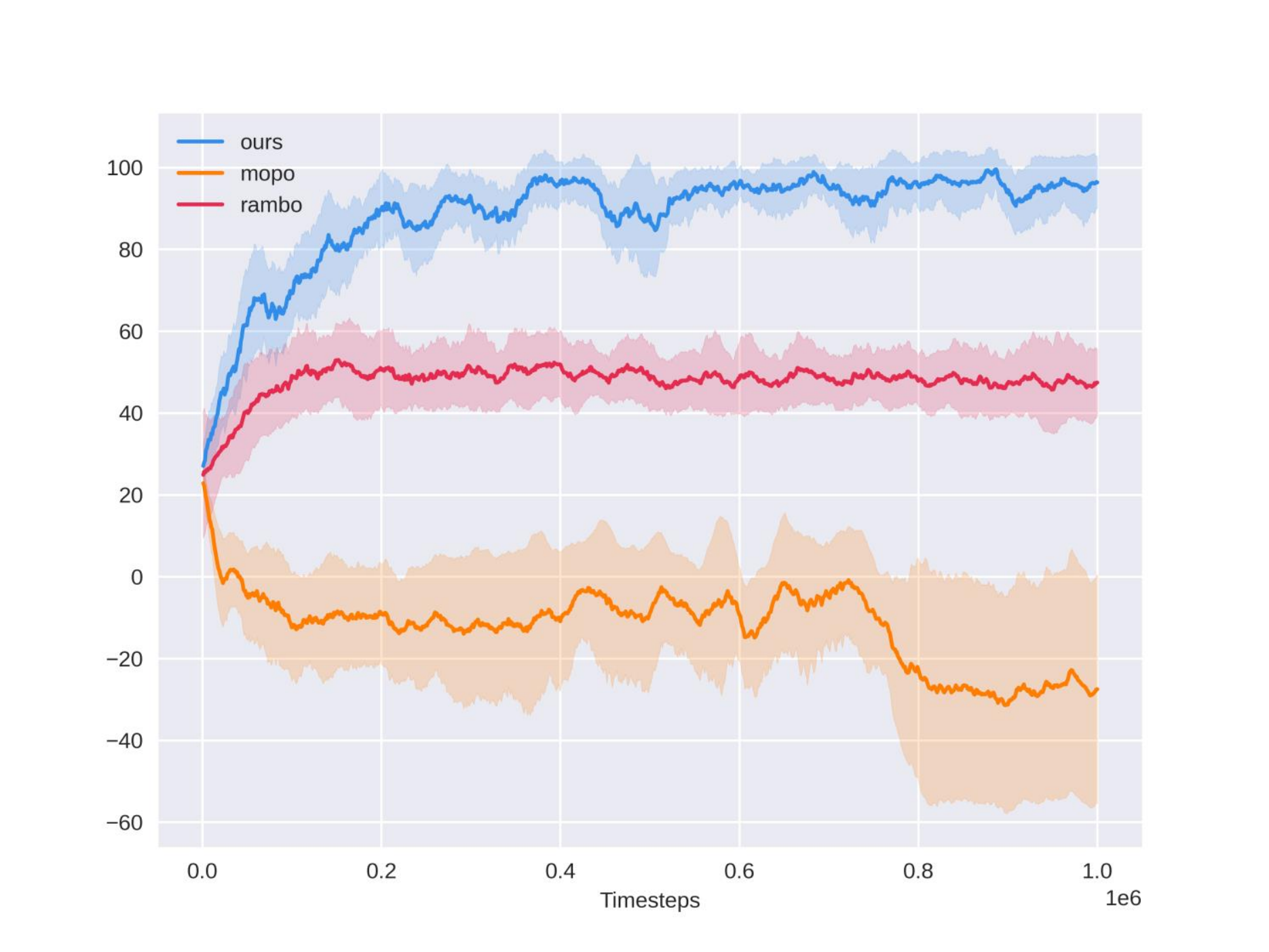}}
        \subfloat[ant-medium]{
		\includegraphics[width=0.5\columnwidth]{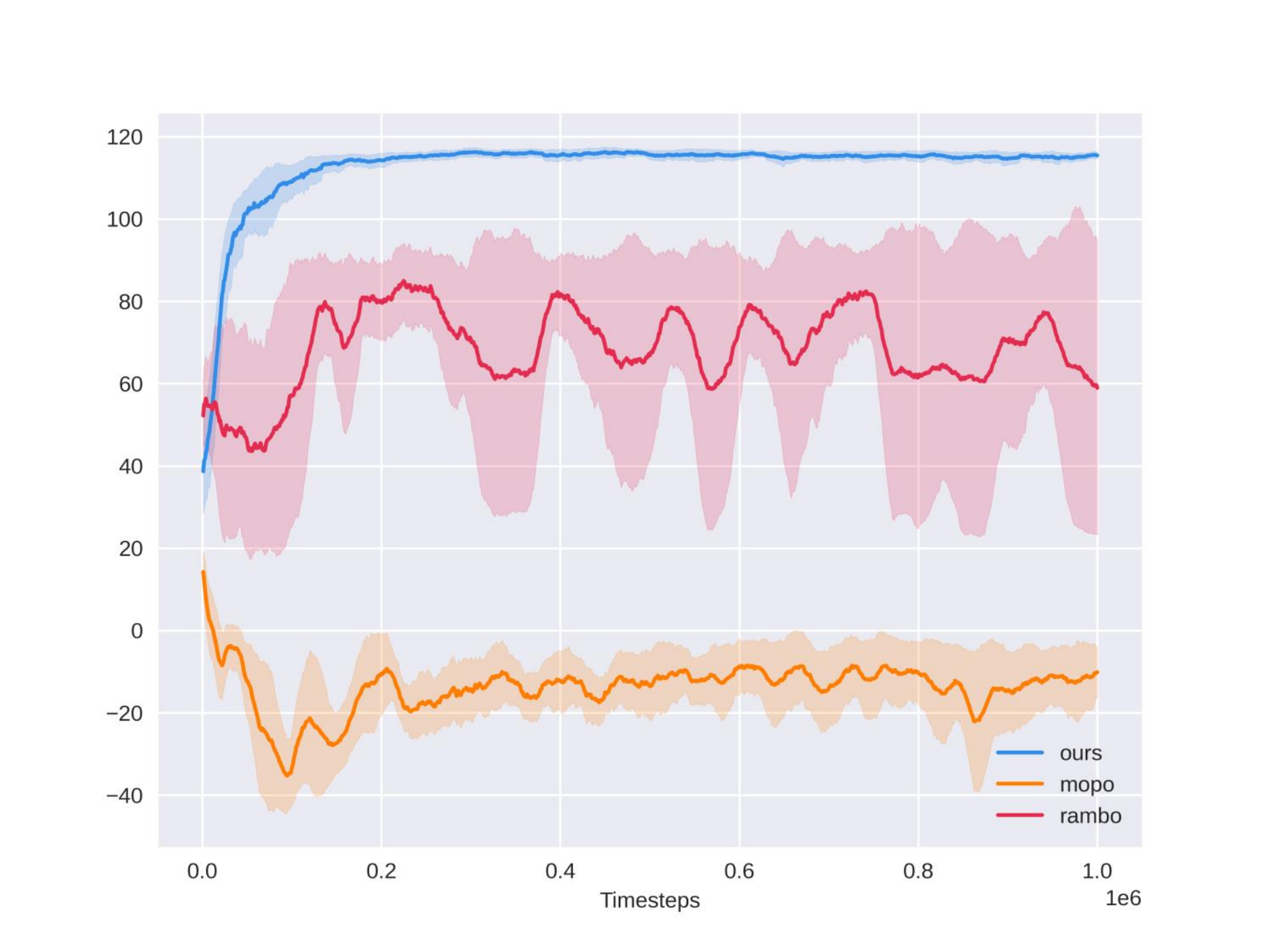}}
	\quad
	\subfloat[halfcheetah-full-replay]{
		\includegraphics[width=0.5\columnwidth]{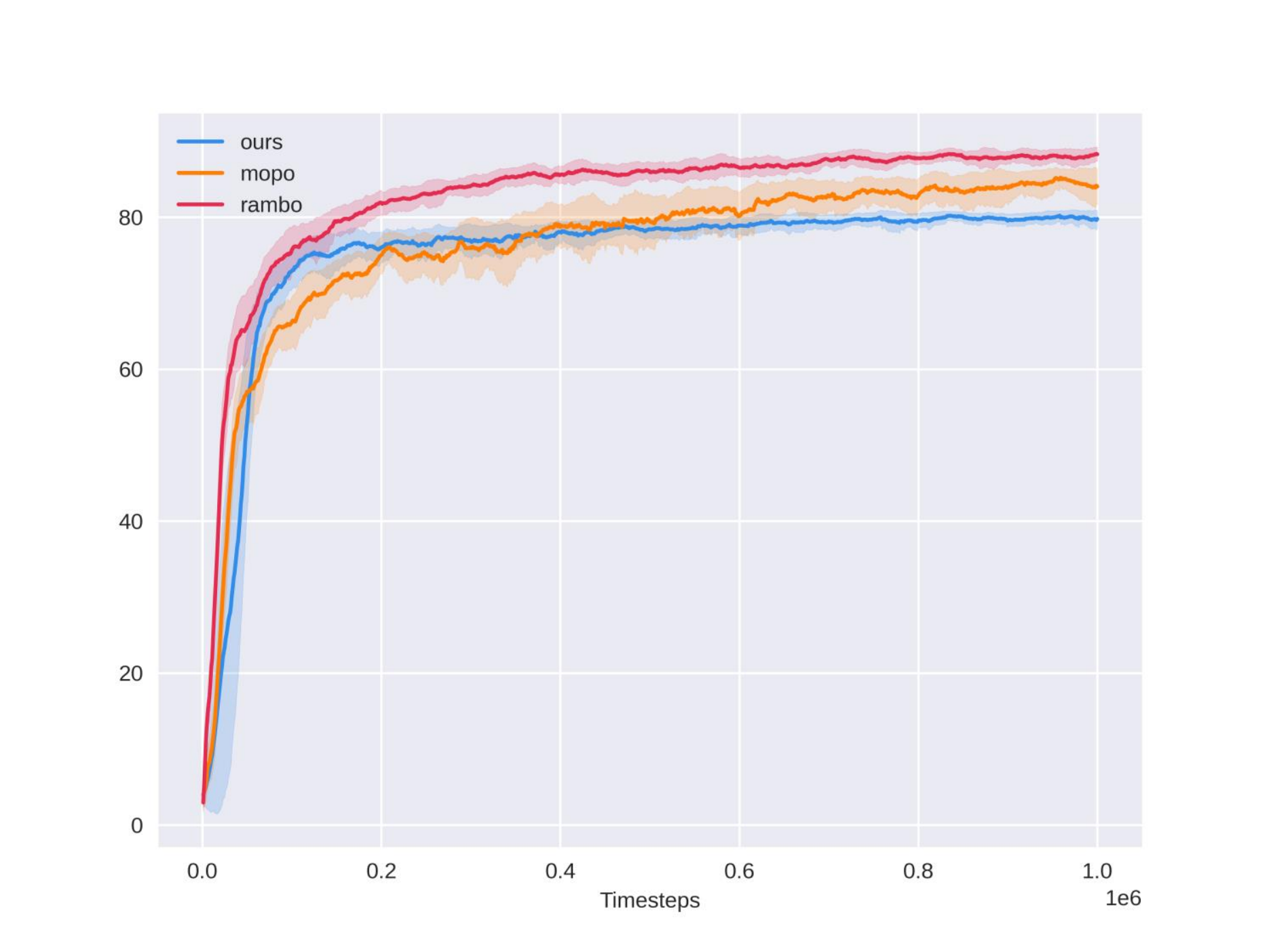}}
	\subfloat[halfcheetah-medium-expert]{
		\includegraphics[width=0.5\columnwidth]{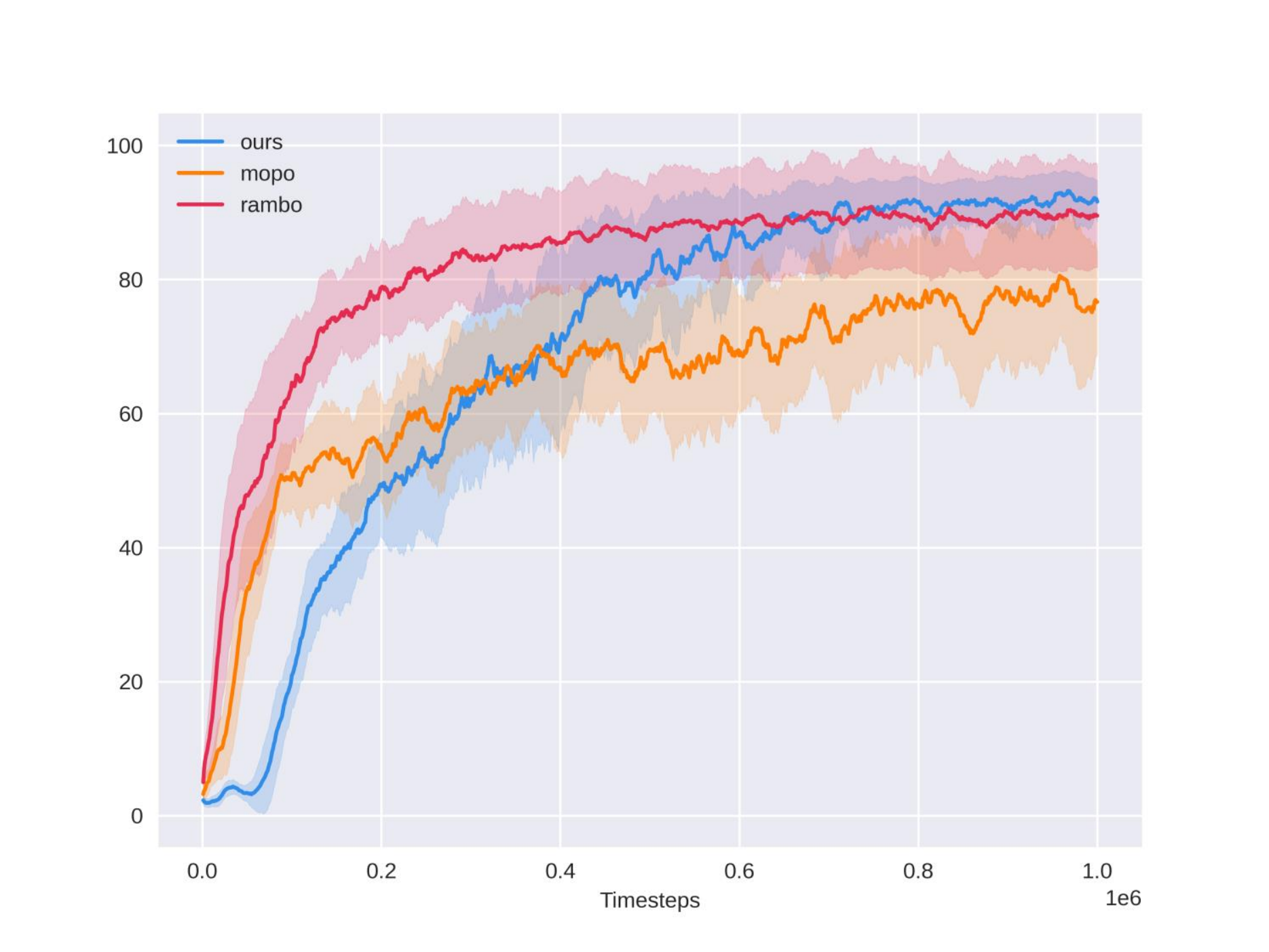}}
	\subfloat[halfcheetah-medium-replay]{
		\includegraphics[width=0.5\columnwidth]{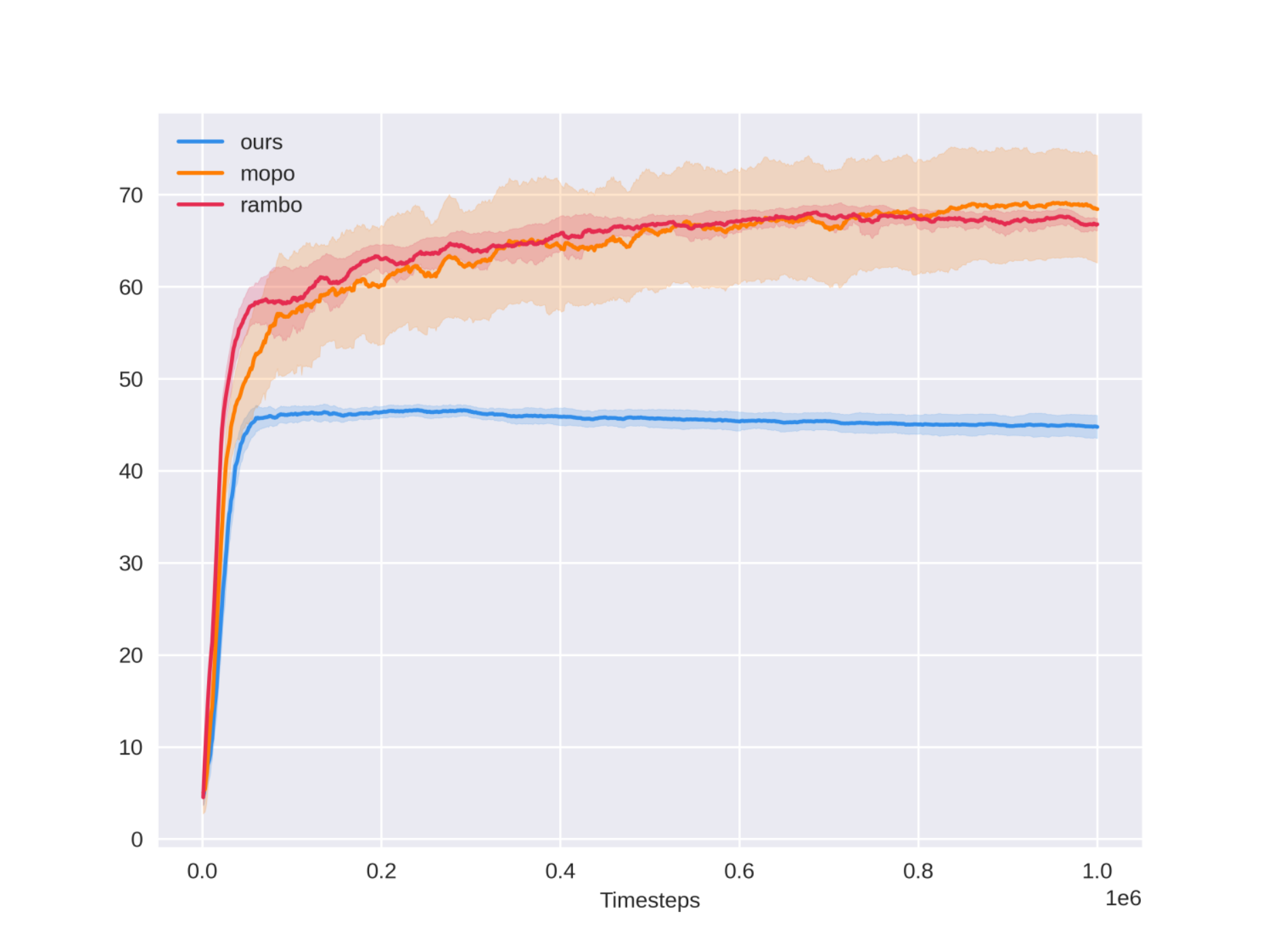}}
        \subfloat[halfcheetah-medium]{
		\includegraphics[width=0.5\columnwidth]{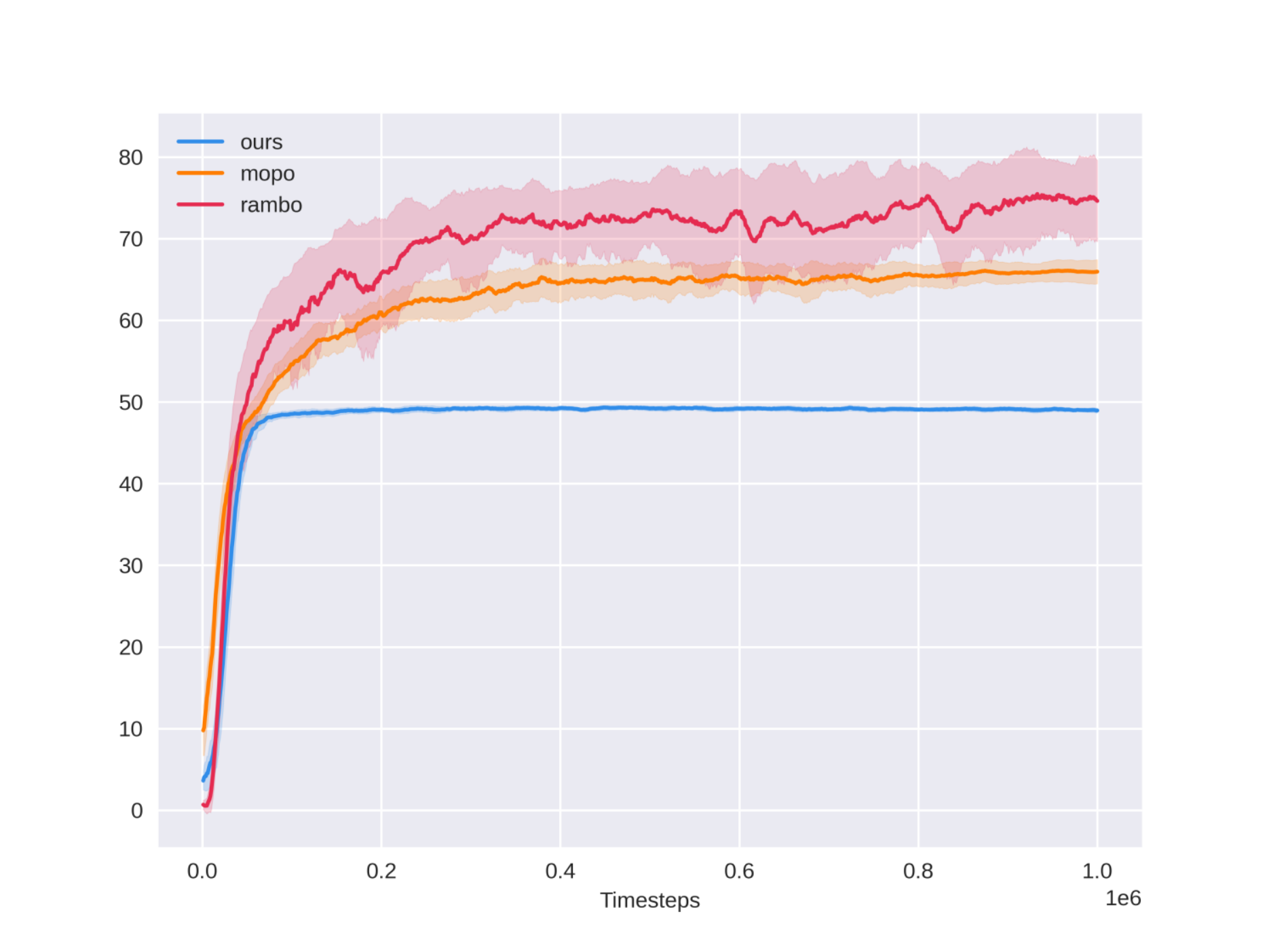}}
	\quad
        \subfloat[hopper-full-replay]{
		\includegraphics[width=0.5\columnwidth]{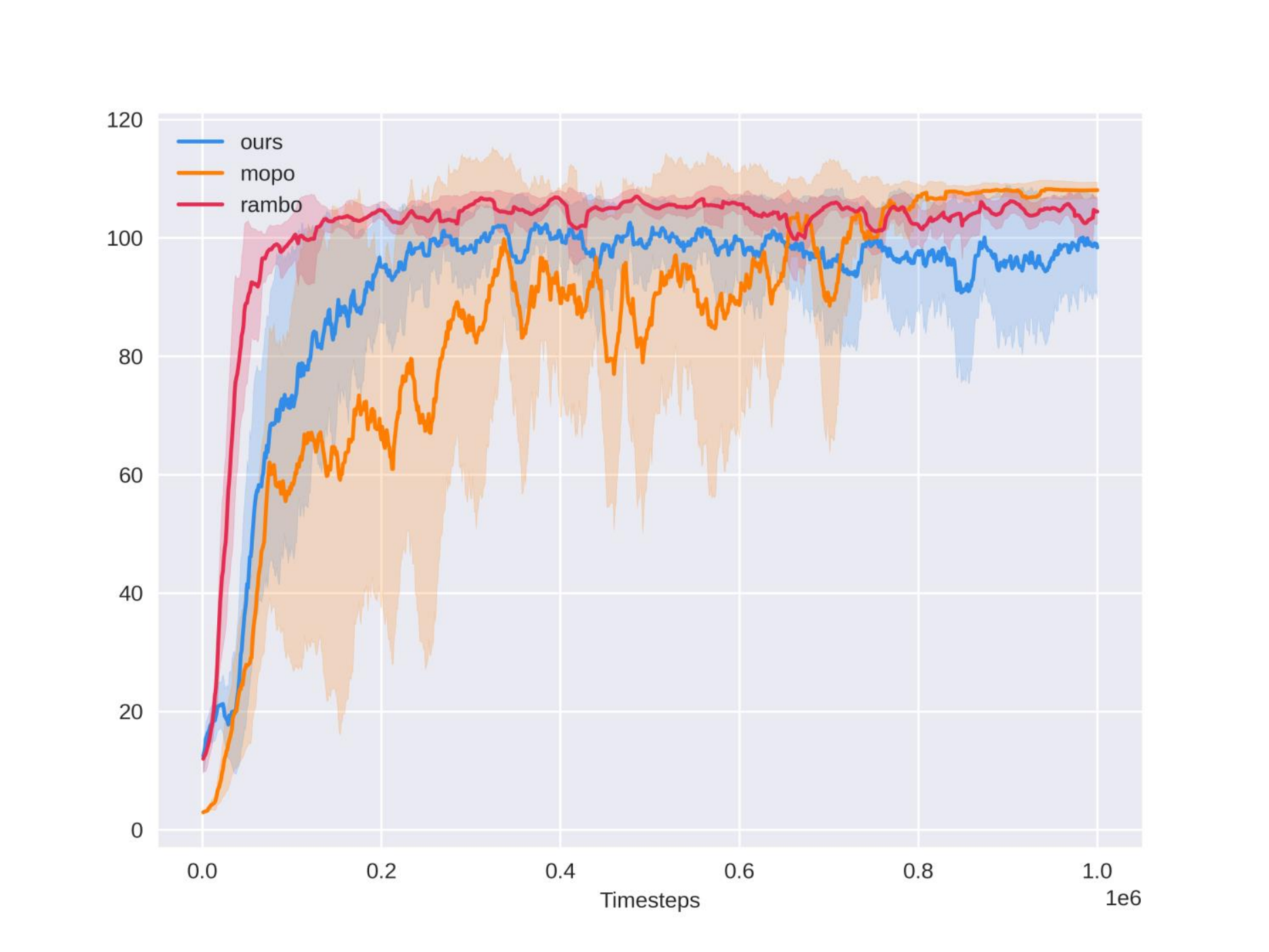}}
	\subfloat[hopper-medium-expert]{
		\includegraphics[width=0.5\columnwidth]{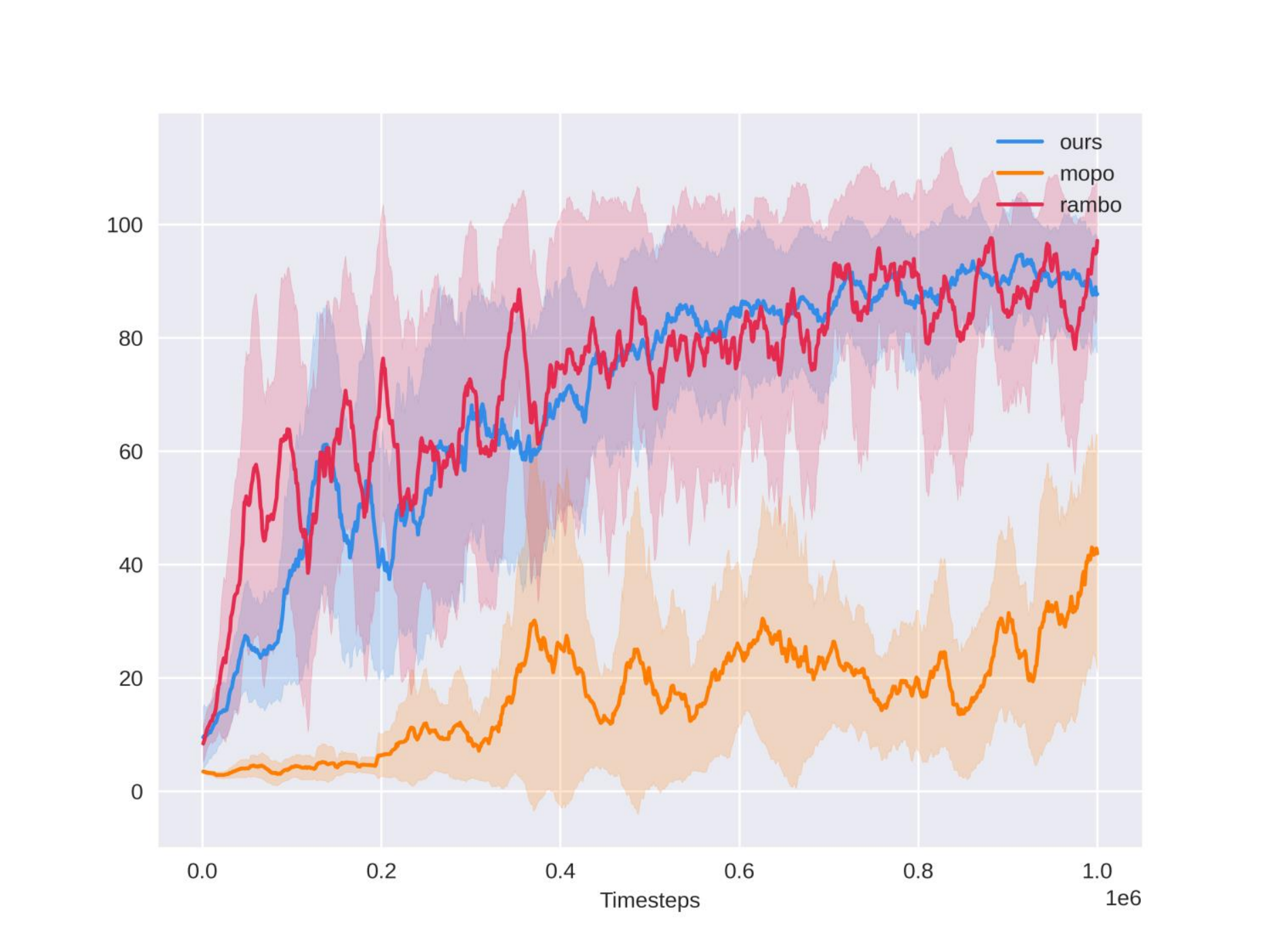}}
	\subfloat[hopper-medium-replay]{
		\includegraphics[width=0.5\columnwidth]{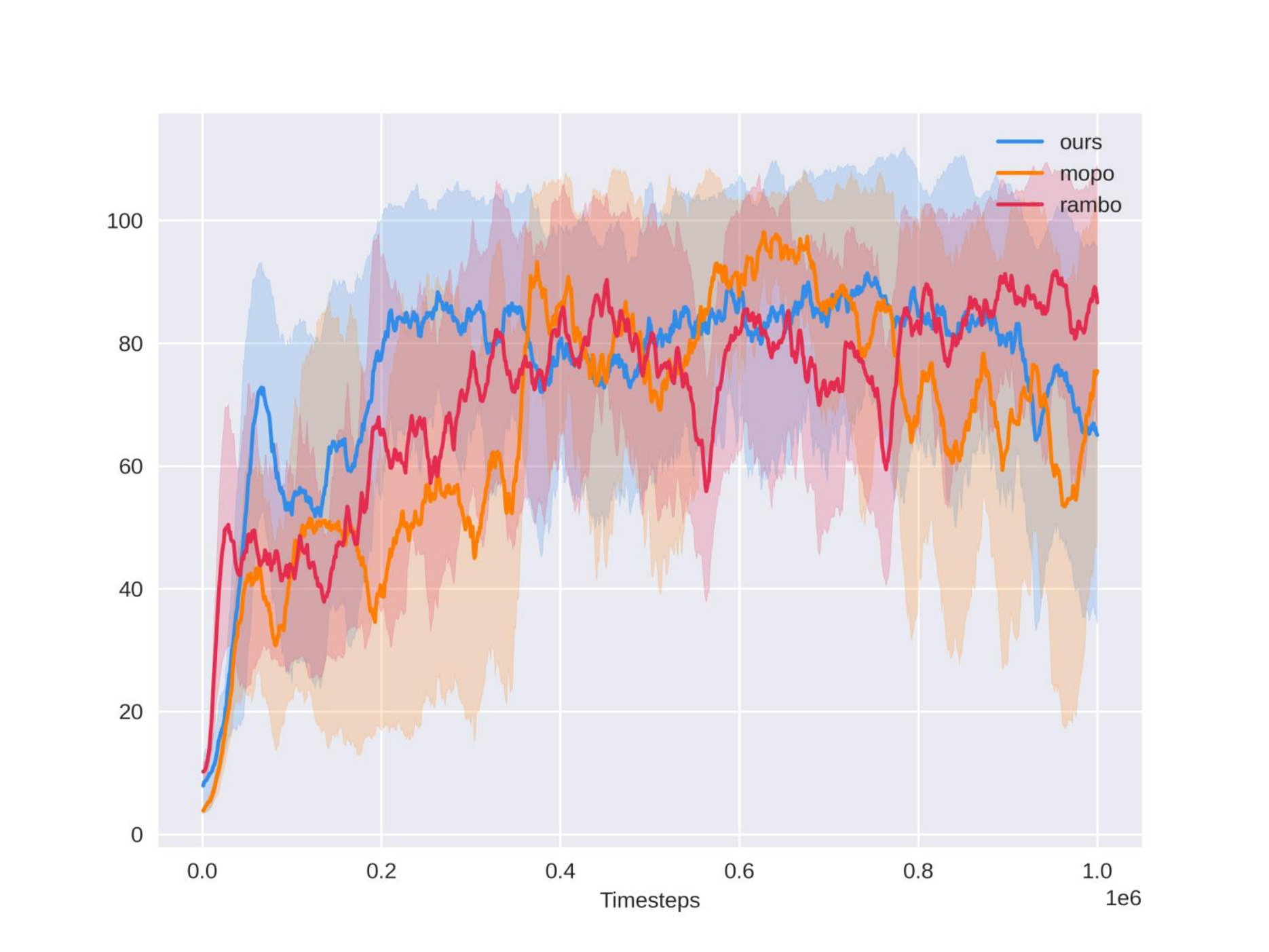}}
        \subfloat[hopper-medium]{
		\includegraphics[width=0.5\columnwidth]{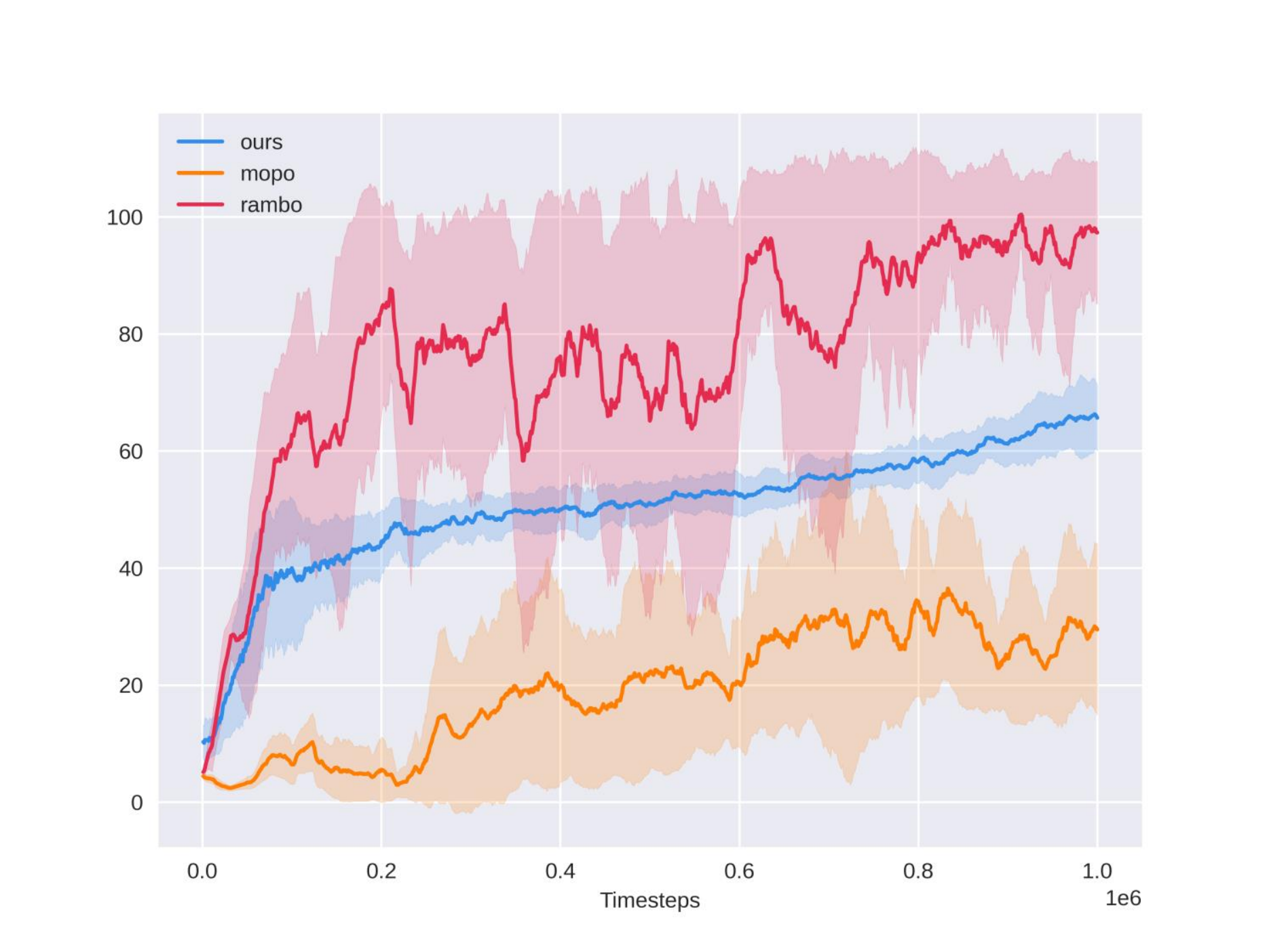}}
	\quad
        \subfloat[walker2d-full-replay]{
		\includegraphics[width=0.5\columnwidth]{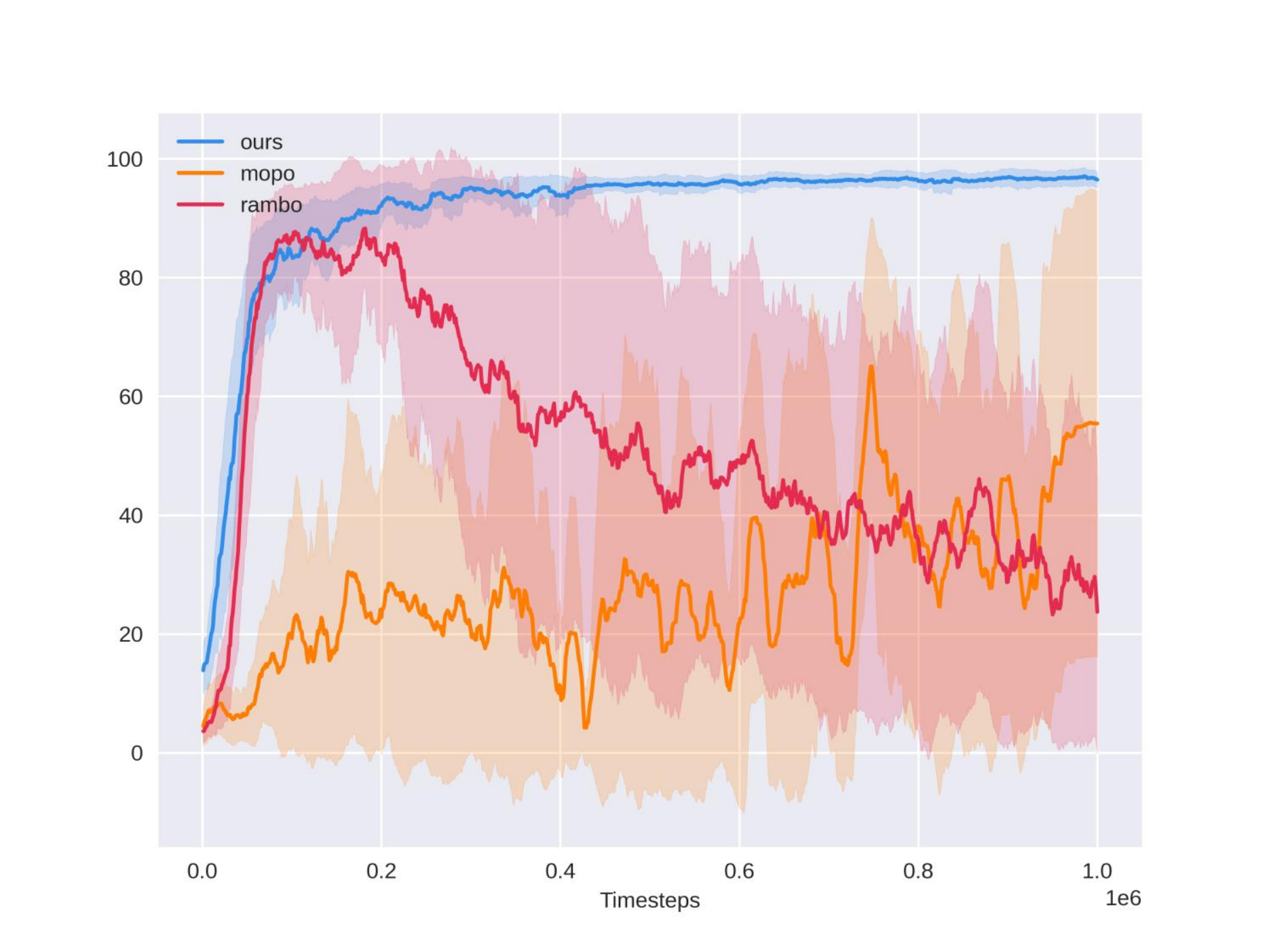}}
	\subfloat[walker2d-medium-expert]{
		\includegraphics[width=0.5\columnwidth]{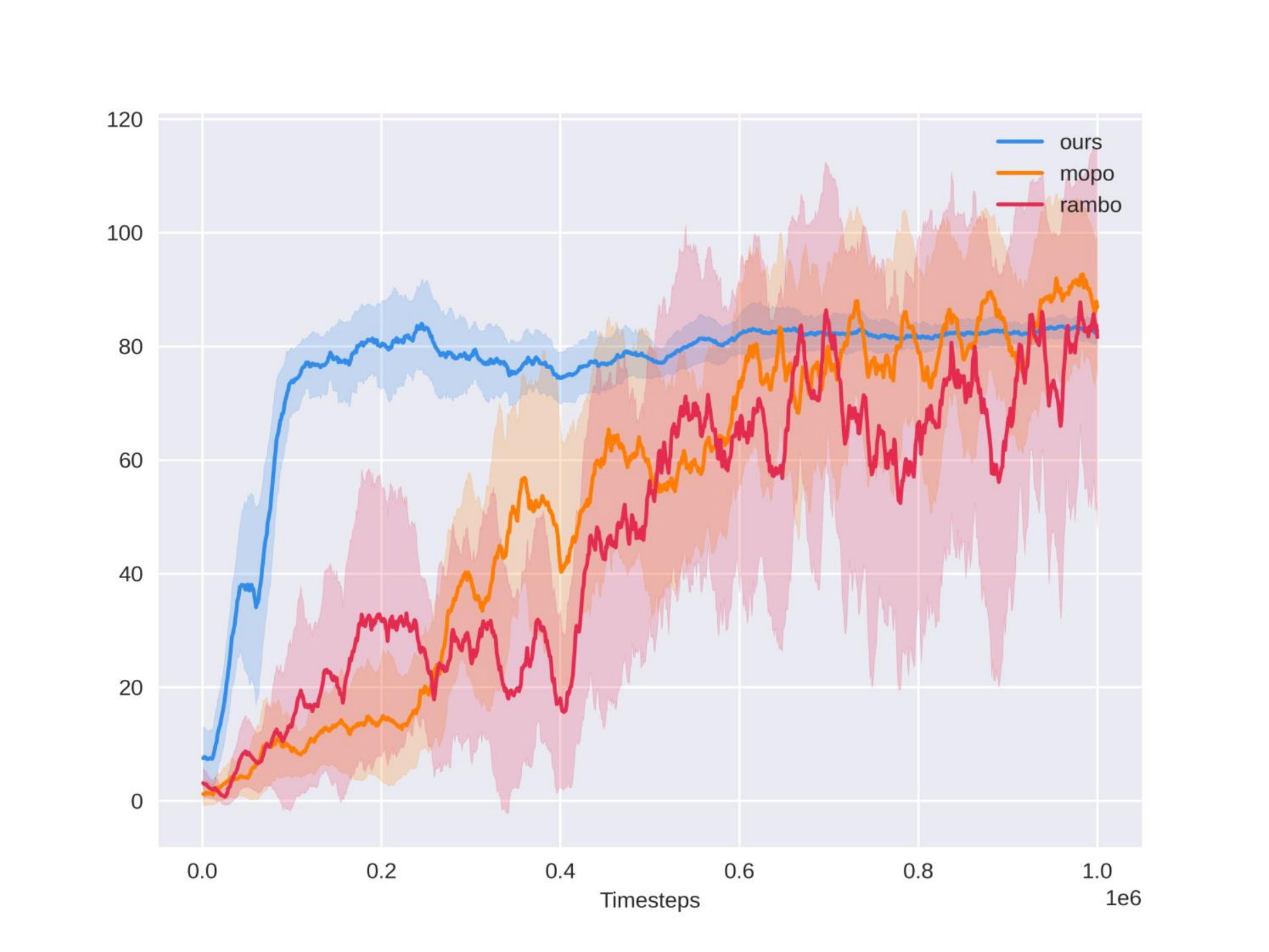}}
	\subfloat[walker2d-medium-replay]{
		\includegraphics[width=0.5\columnwidth]{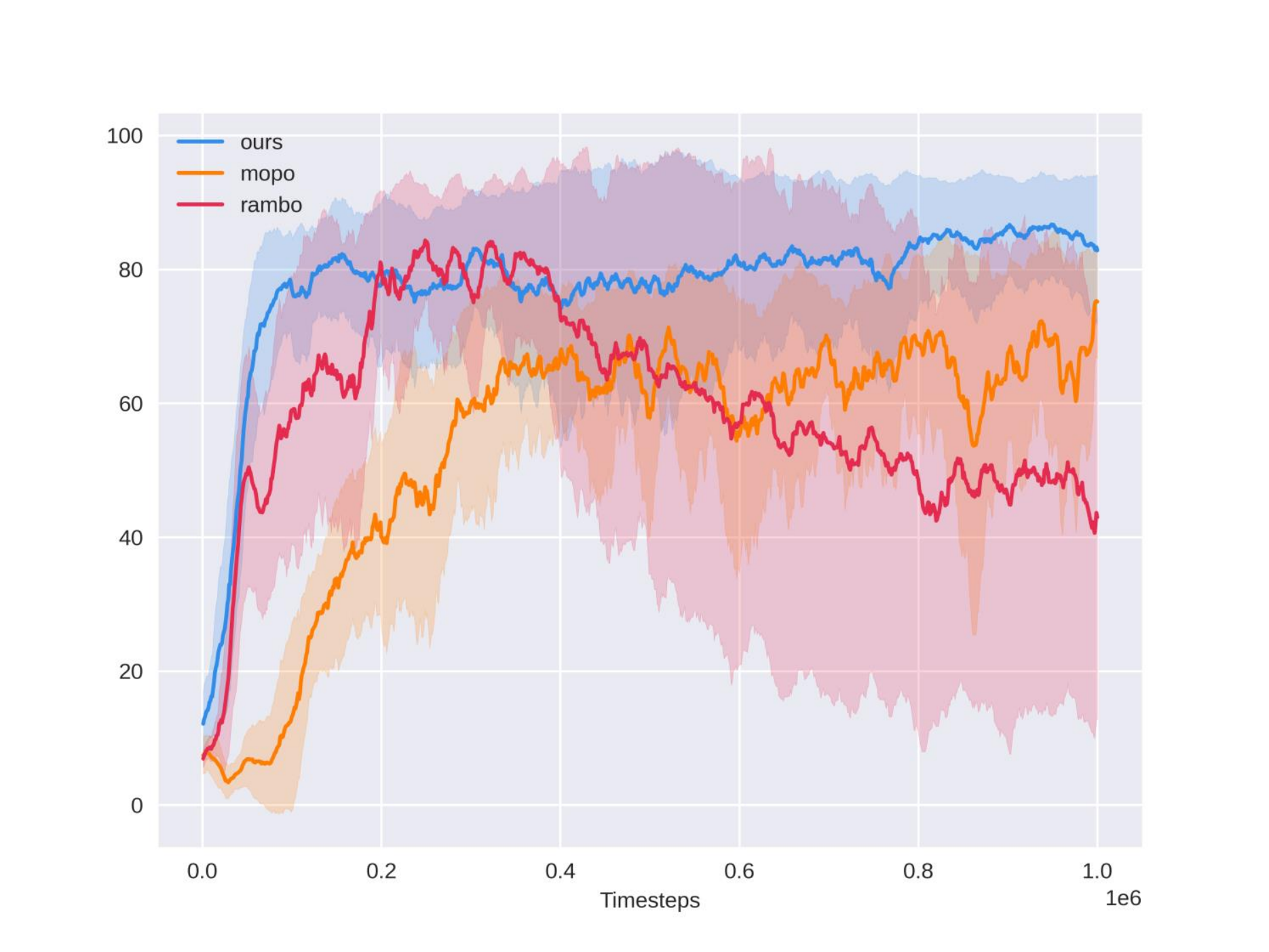}}
        \subfloat[walker2d-medium]{
		\includegraphics[width=0.5\columnwidth]{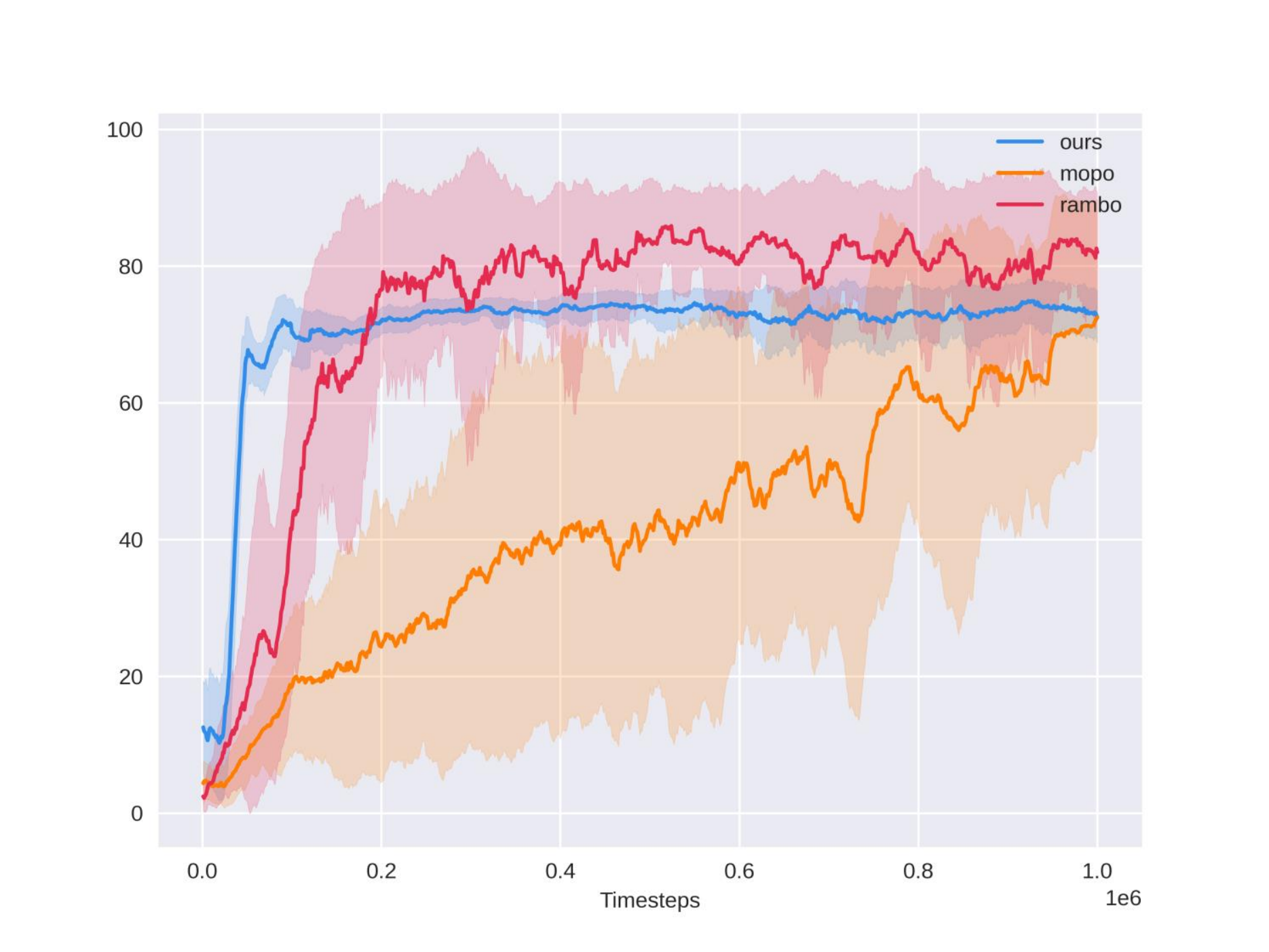}}
	\caption{The episode reward curves during policy learning with three random seeds.}
	\label{scores}
	\vspace{-0.1cm}
\end{figure*}

We visualize the episode reward curves during $1e6$ policy learning, compared with SOTA model-based offline RL algorithms MOPO and RAMBO, as shown in Fig. \ref{scores}. Due to the complexity of the Ant environment, MOPO and RAMBO can not accurately model the environment, which leads to the failure to achieve stable policy improvement. However, the strategies trained by our approach all meet or far exceed expert performance, which demonstrates that Environment Transformer's simulated rollout quality is much greater than the ensemble dynamics models of MOPO and RAMBO. Our approach shows the highest or similar performance with the lowest variance and fastest converging speed in most tasks, which proves our sample efficiency.
Our method performs less well on medium datasets. The main reason is that conservative Q-Learning suffers from the lack of action diversity in medium datasets. As a result, learning a policy that generalizes well becomes more difficult.

\subsection{Long-term Rollout Evaluation}

\begin{figure*}[htbp]
	\centering

        \subfloat[Long-term evaluation in offline datasets]{
		\includegraphics[width=1.0\columnwidth]{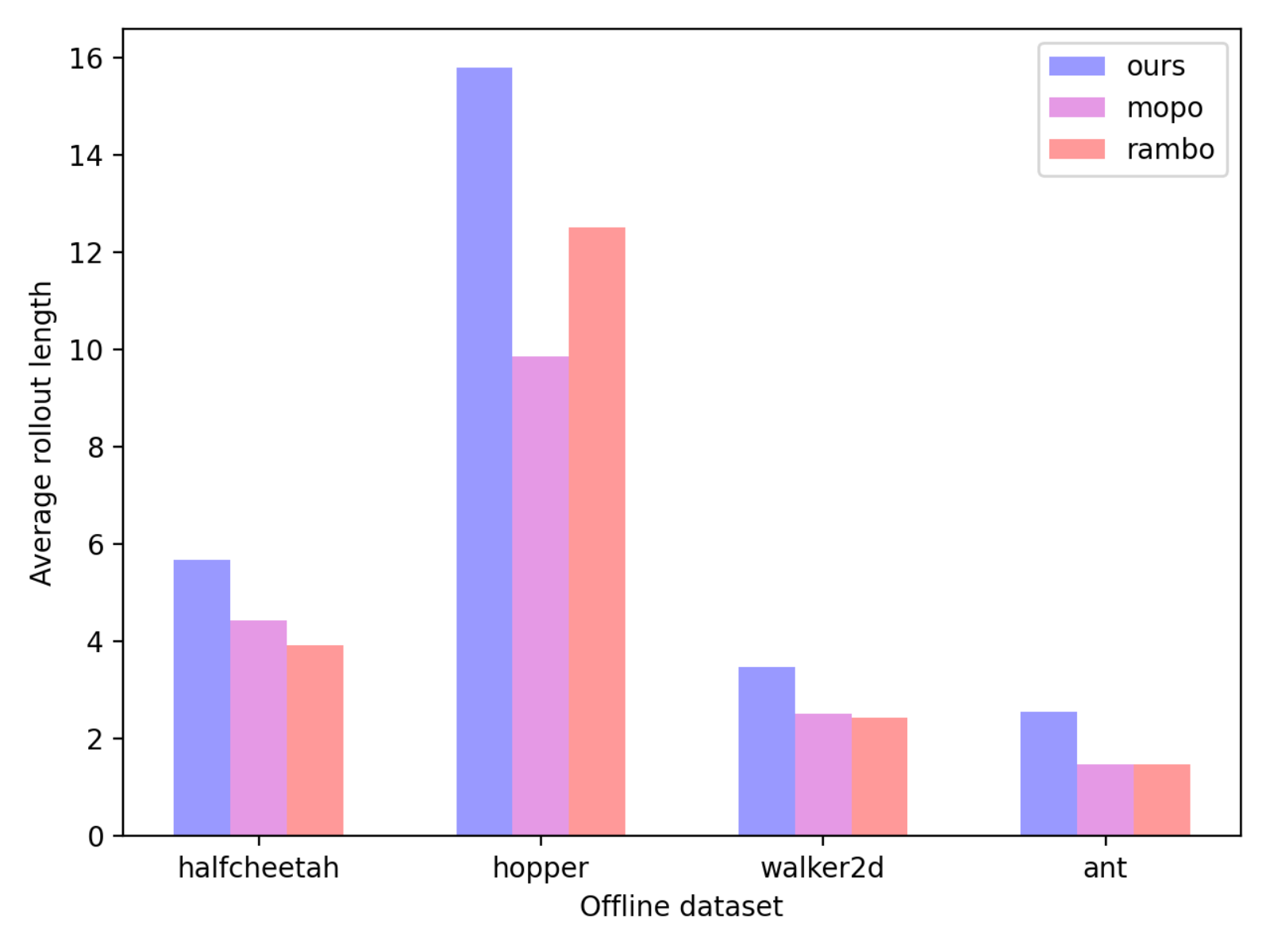}
		\label{fig:long_sequence1}
	}	
	\subfloat[Long-term evaluation in online environments]{
		\includegraphics[width=1.0\columnwidth]{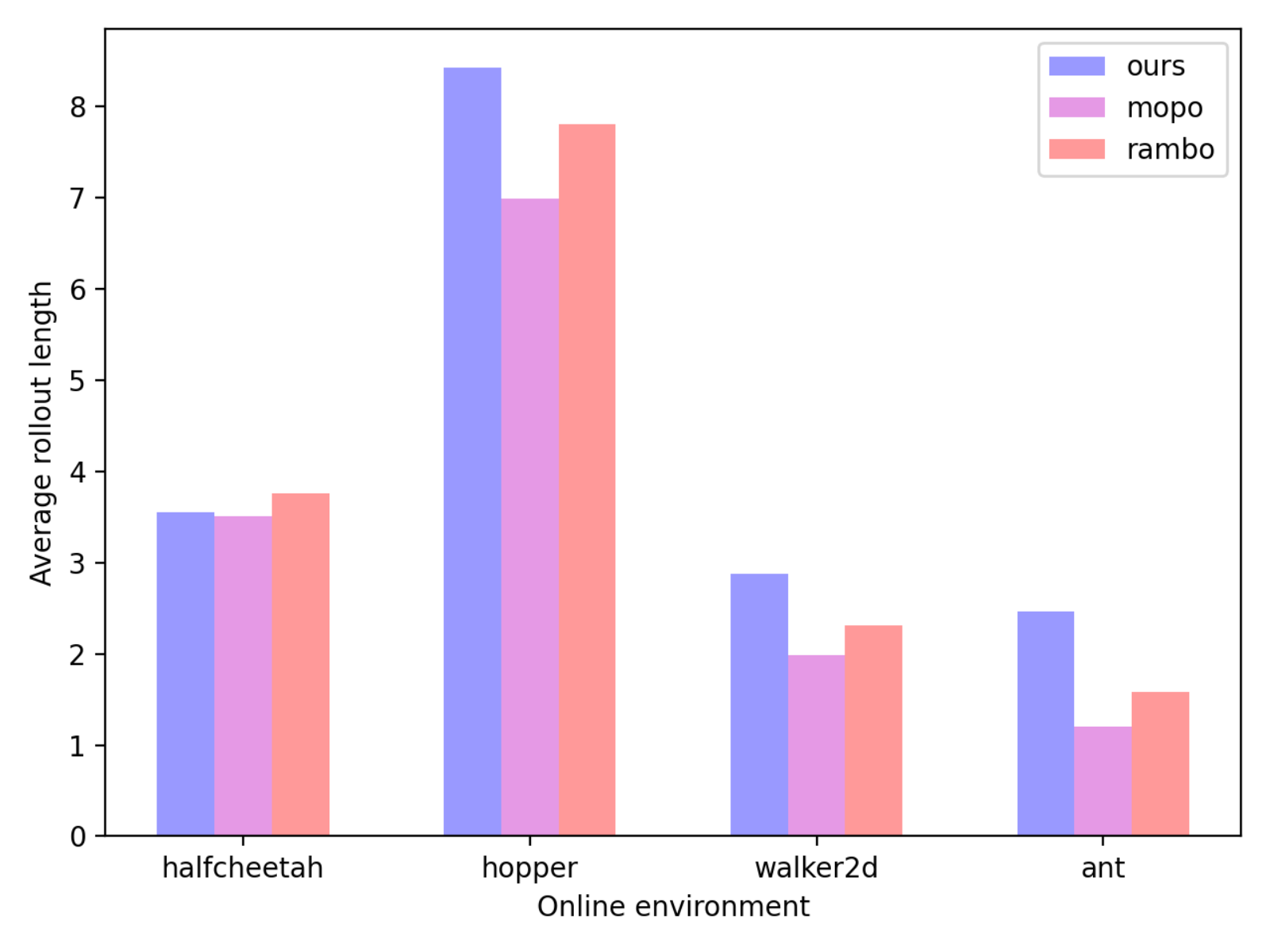}
		\label{fig:long_sequence2}
	}
        
	\caption{The long-term rollout evaluation using both offline datasets and online environments.
	}	
	\label{fig:long_sequence}
\end{figure*}

We design offline and online experiments to evaluate the long-term rollout generation capability for Environment Transformer and dynamics models of MOPO and RAMBO.
In the offline setting, we random sample a long-duration (usually one thousand steps) trajectory from the datasets and then perform rollout simulation for each state-action pair on the trajectory. We record the rollout length when the mse between the normalized predicted state-reward pairs and ground-truth is greater than a threshold. We repeat the above steps hundreds of times and calculate the average for all datasets within the same environment. 
In the online setting, we run random, medium, and expert policies respectively to collect transitions. Similarly, we perform rollout simulation for each state-action pair, update the online environment and calculate the mse between the normalized predictions and ground-truth. We repeat the rollout simulation procedure until the mse is greater than a threshold and record the episode length. At last, we calculate the average for different policies under the same environment.
The threshold is 0.01 for both offline and online settings.
The results of the long-term rollout evaluation are shown in Fig. \ref{fig:long_sequence}.
The average rollout length of our method exceeds MOPO and RAMBO in 7 of 8 environments, which demonstrates that the long-term rollout simulation ability of Environment Transformer is superior to previous model-based offline RL methods.

\section{Conclusions}

In this paper, we study model-based offline RL approaches. Existing model-based offline RL works utilize probabilistic ensemble NN to capture aleatoric and epistemic uncertainty, which leads to an exponential increase in training time and computing resource requirements. Furthermore, these methods suffer from the accumulative errors when simulating long-duration rollouts. In order to solve the above issues, we propose Environment Transformer, an uncertainty-aware sequence modeling architecture. It models the probability distribution of the transition dynamics and reward function to capture aleatoric uncertainty and considers epistemic uncertainty as a learnable noise parameter. Thanks to the accurate modeling of environment dynamics and reward function, it can be combined with arbitrary planning, dynamic programming, or policy learning methods. In this case, We conduct Conservative Q-Learning (CQL) to learn a conservative Q-function based on Environment Transformer for offline RL tasks.

We design simulation experiments on widely studied offline RL benchmarks to evaluate the performance. The experimental results show that our method achieves or exceeds SOTA performance of both model-based and model-free offline RL algorithms. Moreover, we demonstrate that Environment Transformer's simulated rollout quality, sample efficiency, and long-term rollout simulation capability are superior to those of previous model-based offline RL methods.

Environment Transformer provides a new paradigm for merging offline RL and robot tasks, as it offers a low-cost and high-efficiency data acquisition approach for training real-world robots. we believe that it has broad applicability potential. The limitation of our method is insufficient real-world tests. In the future, we intend to optimize this work further and implement more difficult real-world evaluations.





\section*{Acknowledgment}

The authors would like to thank Prof. Barto, Michael Janner, and Costa Huang for insightful discussions. 

This study is supported by the National Natural Science Foundation of China under Grant 52302379, Guangzhou Basic and Applied Basic Research Project 2023A03J0106, Guangdong Province General Universities Youth Innovative Talents Project under Grant 2023KQNCX100, and Guangzhou Municipal Science and Technology Project 2023A03J0011.

\end{document}